\documentclass[letterpaper, 10 pt, conference]{ieeeconf}  

\IEEEoverridecommandlockouts                              

\usepackage{color}
\usepackage{transparent}
\usepackage{import}
\usepackage{multirow}
\usepackage{balance}
\usepackage{graphicx,xcolor}
\usepackage{psfrag}
\usepackage{amsmath,amsfonts}

\usepackage{amssymb}  
\usepackage{makecell}

\usepackage{booktabs,tabularx,caption,ragged2e}
\usepackage{cite}
\usepackage{setspace}
\usepackage{times,xspace}
\usepackage[font={footnotesize}]{caption}
\usepackage{makecell}
\usepackage{algorithm} 
\usepackage{algpseudocode} 
\usepackage{nicefrac}
\usepackage{romannum}
\usepackage{subcaption}
\usepackage{siunitx}
\usepackage{url}
\usepackage[bottom]{footmisc}

\usepackage[makeroom]{cancel}
\usepackage{comment}

\usepackage{mathtools}

\usepackage{array}
\usepackage{stmaryrd}
\usepackage{accents}
\usepackage{ntheorem}
\theoremstyle{break}
\theoremheaderfont{\normalfont\bfseries}
\theoremseparator{:}

\RequirePackage[                    
strict=true,                    
style=german                    
]{csquotes}

\definecolor{RPTHred}{RGB}{227, 27, 35}
\definecolor{TUMblue}{RGB}{0, 101, 189}

\usepackage{pifont}

\usepackage[symbols, toc,nonumberlist, sanitizesort]{glossaries}
\usepackage{color}
\usepackage{seqsplit}
\usepackage{xargs}
\usepackage{xstring}

\newglossary[glignoredl]{ignored}{glignored}{glignoredin}{Ignored Glossary}

\newglossary{time}{gls1}{glo1}{Time}
\newglossary{longitudinalplanning}{gls2}{glo2}{Longitudinal Trajectory Planning}
\newglossary{lateralplanning}{gls3}{glo3}{Lateral Trajectory Planning}
\newglossary{cosy}{gls4}{glo4}{Coordinate Systems}
\newglossary{misc}{gls5}{glo5}{Miscellaneous}
\newglossary{state}{gls6}{glo6}{State Variables of the Ego Vehicle}
\newglossary{refpath}{gls7}{glo7}{Reference Path}
\newglossary{shape}{gls8}{glo8}{Ego Vehicle Shape}
\newglossary{obs}{gls9}{glo9}{Obstacles}
\newglossary{reach}{gls10}{glo10}{Reachability Analysis}
\newglossary{drivingcorridor}{gls11}{glo11}{Driving Corridor Identification}
\newglossary{constraints}{gls12}{glo12}{Collision Avoidance Constraints}
\newglossary{trajectoryrepairing}{gls13}{glo13}{Trajectory Repairing}
\newglossary{signaltemporallogic}{gls14}{glo14}{Signal Temporal Logic}

\makenoidxglossaries
\glsnoexpandfields

\newcommand{\shorteq}{%
	\settowidth{\@tempdima}{-}%
	\resizebox{\@tempdima}{\height}{=}%
}

\newcommand{\ndec}{\glslink{ndec}{\alpha}}
\newglossaryentry{ndec}{
	name={\ensuremath{\ndec}},
	description={Number of tactical decisions},
	sort=5 a,
	type=misc
}

\newcommand{\nobs}{\glslink{nobs}{\sigma}}
\newglossaryentry{nobs}{
	name={\ensuremath{\nobs}},
	description={Number of obstacles},
	sort=5 b,
	type=misc
}

\newcommand{\cartesian}{\mathtt{G}}
\newcommand{\local}{\mathtt{L}}
\newcommand{\vehiclefixed}{\mathtt{V}}
\newcommandx{\cosy}[1]{F^{#1}}
\newcommand{\transformation}{T}

\newcommand{\curvframe}{\glslink{curvframe}{\cosy{\local}}}
\newglossaryentry{curvframe}{
	name={\ensuremath{\curvframe}},
	description={Local, curvilinear coordinate frame},
	sort=b,
	type=cosy
}

\newcommandx{\localvar}[1]{\glslink{localvar}{\leftidx{^{\local}}{#1}}}
\newglossaryentry{localvar}{
	name={\ensuremath{\localvar{\square}}},
	description={Variable $\square$ in $\curvframe$},
	parent=curvframe,
	sort=b,
	type=cosy
}

\newcommand{\cartesianframe}{\glslink{cartesianframe}{\cosy{\cartesian}}}
\newglossaryentry{cartesianframe}{
	name={\ensuremath{\cartesianframe}},
	description={Global, Cartesian coordinate frame},
	sort=b,
	type=cosy
}

\newcommandx{\globalvar}[1]{\glslink{globalvar}{\leftidx{^{\cartesian}}{#1}}}
\newglossaryentry{globalvar}{
	name={\ensuremath{\globalvar{\square}}},
	description={Variable $\square$ in $\cartesianframe$},
	parent=cartesianframe,
	sort=b,
	type=cosy
}

\newcommand{\vehicleframe}{\glslink{vehicleframe}{\cosy{\vehiclefixed}}}
\newglossaryentry{vehicleframe}{
	name={\ensuremath{\vehicleframe}},
	description={Vehicle-fixed coordinate frame},
	sort=b,
	type=cosy
}

\newcommand{\curvframelondir}{\glslink{curvframelondir}{\zeta}}
\newglossaryentry{curvframelondir}{
	name={\ensuremath{\curvframelondir}},
	description={Longitudinal direction in $\curvframe$},
	parent=curvframe,
	sort=b,
	type=cosy
}

\newcommand{\curvframelatdir}{\glslink{curvframelatdir}{\eta}}
\newglossaryentry{curvframelatdir}{
	name={\ensuremath{\curvframelatdir}},
	description={Lateral direction in $\curvframe$},
	parent=curvframe,
	sort=b,
	type=cosy
}

\newcommandx{\transformationcurvcartesian}[1][1={\slon}, usedefault]{\glslink{transformationcurvcartesian}{\transformation_{\local}^{\cartesian}\left(#1\right)}}
\newglossaryentry{transformationcurvcartesian}{
	name={\ensuremath{\transformationcurvcartesian}},
	description={Transformation from $\curvframe$ to $\cartesianframe$},
	sort=b b,
	type=cosy
}

\newcommandx{\transformationvehiclecartesian}[1][1=\orientation, usedefault]{\glslink{transformationvehiclecartesian}{\transformation_{\vehiclefixed}^{\cartesian}\left(#1\right)}}
\newglossaryentry{transformationvehiclecartesian}{
	name={\ensuremath{\transformationvehiclecartesian}},
	description={Transformation from $\vehicleframe$ to $\cartesianframe$},
	sort=b b,
	type=cosy
}

\newcommand{\maxval}[1]{\glslink{maxval}{\overline{#1}}}
\newglossaryentry{maxval}{
	name={\ensuremath{\maxval{\square}}},
	description={Maximum admissible value of a variable $\square$},
	sort=6a,
	type=state
}

\newcommand{\minval}[1]{\glslink{minval}{\underline{#1}}}
\newglossaryentry{minval}{
	name={\ensuremath{\minval{\square}}},
	description={Minimum admissible value of a variable $\square$},
	sort=6b,
	type=state
}

\newcommand{\pos}{s}
\newcommand{\velocity}{v}
\newcommand{\acceleration}{a}
\newcommand{\jerk}{j}

\newcommand{\slon}{\glslink{slon}{\pos_{\curvframelondir}}}
\newglossaryentry{slon}{
	name={\ensuremath{\slon}},
	description={Longitudinal position in $\curvframe$},
	sort=6c,
	type=state
}
\newcommandx{\slonk}[1][1=\timestep, usedefault]{
	\glslink{slonk}{\pos_{{\curvframelondir, #1}}}}
\newglossaryentry{slonk}{
	name={\ensuremath{\slonk}},
	description={Longitudinal position $\slon$ at time step $\timestep$},
	sort=6c,
	parent=slon,
	type=state
}

\newcommand{\dslon}{\glslink{dslon}{\dot{\pos}_{\curvframelondir}}}
\newglossaryentry{dslon}{
	name={\ensuremath{\dslon}},
	description={Time derivative of longitudinal position in $\curvframe$},
	type=ignored,
}

\newcommand{\sminlon}{\glslink{sminlon}{\minval{\pos}_{\curvframelondir}}}
\newglossaryentry{sminlon}{
	name={\ensuremath{\sminlon}},
	description={Minimum value of $\slon$},
	sort=6c,
	type=state,
	parent=slon
}

\newcommandx{\sminlonk}[1][1=\timestep, usedefault]{\glslink{sminlonk}{\minval{\pos}_{\curvframelondir, #1}}}
\newglossaryentry{sminlonk}{
	name={\ensuremath{\sminlonk}},
	description={Minimum value of $\slon$ at time step $\timestep$},
	sort=6c,
	type=state,
	parent=sminlon
}

\newcommand{\smaxlon}{\glslink{smaxlon}{\maxval{\pos}_{\curvframelondir}}}
\newglossaryentry{smaxlon}{
	name={\ensuremath{\smaxlon}},
	description={Maximum value of $\slon$},
	sort=6c,
	type=state,
	parent=slon
}

\newcommandx{\smaxlonk}[1][1=\timestep, usedefault]{\glslink{smaxlonk}{\maxval{\pos}_{\curvframelondir, #1}}}
\newglossaryentry{smaxlonk}{
	name={\ensuremath{\smaxlonk}},
	description={Maximum value of $\slon$ at time step $\timestep$},
	sort=6c,
	type=state,
	parent=smaxlon
}

\newcommand{\vlon}{\glslink{vlon}{\velocity_{\curvframelondir}}}
\newglossaryentry{vlon}{
	name={\ensuremath{\vlon}},
	description={Longitudinal velocity in $\curvframe$},
	sort=6d,
	type=state
}
\newcommandx{\vlonk}[1][1=\timestep, usedefault]{\glslink{vlonk}{\velocity_{\curvframelondir, #1}}}
\newglossaryentry{vlonk}{
	name={\ensuremath{\vlonk}},
	description={Longitudinal velocity $\vlon$ at time step $\timestep$},
	sort=6d,
	type=state,
	parent=vlon
}

\newcommand{\vminlon}{\glslink{vminlon}{\minval{\velocity}_{\curvframelondir}}}
\newglossaryentry{vminlon}{
	name={\ensuremath{\vminlon}},
	description={Minimum value of $\vlon$},
	sort=6d,
	type=state,
	parent=vlon
}

\newcommand{\vmaxlon}{\glslink{vmaxlon}{\maxval{\velocity}_{\curvframelondir}}}
\newglossaryentry{vmaxlon}{
	name={\ensuremath{\vmaxlon}},
	description={Maximum value of $\vlon$},
	sort=6d,
	type=state,
	parent=vlon
}

\newcommand{\alon}{\glslink{alon}{\acceleration_{\curvframelondir}}}
\newglossaryentry{alon}{
	name={\ensuremath{\alon}},
	description={Longitudinal acceleration in $\curvframe$},
	sort=6e,
	type=state,
}

\newcommandx{\alonk}[1][1=\timestep, usedefault]{\acceleration_{\curvframelondir, #1}}
\newglossaryentry{alonk}{
	name={\ensuremath{\alonk}},
	description={Longitudinal acceleration $\alon$ at time step $\timestep$},
	sort=6e,
	type=state,
	parent=alon
}

\newcommand{\aminlon}{\glslink{aminlon}{\minval{\acceleration}_{\curvframelondir}}}
\newglossaryentry{aminlon}{
	name={\ensuremath{\aminlon}},
	description={Minimum value of $\alon$},
	sort=6e,
	type=state,
	parent=alon
}

\newcommand{\amaxlon}{\glslink{amaxlon}{\maxval{\acceleration}_{\curvframelondir}}}
\newglossaryentry{amaxlon}{
	name={\ensuremath{\amaxlon}},
	description={Maximum value of $\alon$},
	sort=6e,
	type=state,
	parent=alon
}

\newcommand{\jlon}{\glslink{jlon}{\jerk_{\curvframelondir}}}
\newglossaryentry{jlon}{
	name={\ensuremath{\jlon}},
	description={Longitudinal jerk in $\curvframe$},
	sort=6f,
	type=state,
}

\newcommandx{\jlonk}[1][1=\timestep, usedefault]{\glslink{jlonk}{\jerk_{\curvframelondir, #1}}}
\newglossaryentry{jlonk}{
	name={\ensuremath{\jlonk}},
	description={Longitudinal jerk $jlonk$ at time step $\timestep$},
	sort=6f,
	type=state,
	parent=jlon
}

\newcommand{\slat}{\glslink{slat}{\pos_{\curvframelatdir}}}
\newglossaryentry{slat}{
	name={\ensuremath{\slat}},
	description={Lateral position in $\curvframe$},
	sort=6g,
	type=state
}

\newcommandx{\slatk}[1][1=\timestep, usedefault]{
	\glslink{slatk}{\pos_{{\curvframelatdir, #1}}}}
\newglossaryentry{slatk}{
	name={\ensuremath{\slatk}},
	description={Lateral position $\slat$ at time step $\timestep$},
	sort=6g,
	type=state,
	parent=slat
}

\newcommand{\dslat}{\glslink{dslat}{\dot{\pos}_{\curvframelatdir}}}
\newglossaryentry{dslat}{
	name={\ensuremath{\dslat}},
	description={Time derivative of lateral position in $\curvframe$},
	type=ignored
}

\newcommand{\vlat}{\glslink{vlat}{\velocity_{\curvframelatdir}}}
\newglossaryentry{vlat}{
	name={\ensuremath{\vlat}},
	description={Lateral velocity in $\curvframe$},
	sort=6h,
	type=state
}

\newcommandx{\vlatk}[1][1=\timestep, usedefault]{\glslink{vlatk}{\velocity_{\curvframelatdir, #1}}}
\newglossaryentry{vlatk}{
	name={\ensuremath{\vlatk}},
	description={Lateral velocity $\vlat$ at time step $\timestep$},
	sort=6h,
	type=state,
	parent=vlat
}

\newcommand{\vminlat}{\glslink{vminlat}{\minval{\velocity}_{\curvframelatdir}}}
\newglossaryentry{vminlat}{
	name={\ensuremath{\vminlat}},
	description={Minimum value of $\vlat$},
	sort=6h,
	type=state,
	parent=vlat
}

\newcommand{\vmaxlat}{\glslink{vmaxlat}{\maxval{\velocity}_{\curvframelatdir}}}
\newglossaryentry{vmaxlat}{
	name={\ensuremath{\vmaxlat}},
	description={Minimum value of $\vlat$},
	sort=6h,
	type=state,
	parent=vlat
}

\newcommand{\alat}{\glslink{alat}{\acceleration_{\curvframelatdir}}}
\newglossaryentry{alat}{
	name={\ensuremath{\alat}},
	description={Lateral acceleration in $\curvframe$},
	sort=6i,
	type=state
}

\newcommandx{\alatk}[1][1=\timestep, usedefault]{\acceleration_{\curvframelatdir, #1}}
\newglossaryentry{alatk}{
	name={\ensuremath{\alatk}},
	description={Lateral acceleration $\alatk$ at time step $\timestep$},
	sort=6i,
	type=state,
	parent=alat
}

\newcommand{\aminlat}{\glslink{aminlat}{\minval{\acceleration}_{\curvframelatdir}}}
\newglossaryentry{aminlat}{
	name={\ensuremath{\aminlat}},
	description={Minimum value of $\alat$},
	sort=6i,
	type=state,
	parent=alat
}

\newcommand{\amaxlat}{\glslink{amaxlat}{\maxval{\acceleration}_{\curvframelatdir}}}
\newglossaryentry{amaxlat}{
	name={\ensuremath{\amaxlat}},
	description={Maximum value of $\alat$},
	sort=6i,
	type=state,
	parent=alat
}

\newcommand{\sx}{\glslink{sx}{\pos_{x}}}
\newglossaryentry{sx}{
	name={\ensuremath{\sx}},
	description={Longitudinal position in $\cartesianframe$},
	sort=6j,
	type=state
}

\newcommand{\sy}{\glslink{sy}{\pos_{y}}}
\newglossaryentry{sy}{
	name={\ensuremath{\sy}},
	description={Lateral position in $\cartesianframe$},
	sort=6k,
	type=state
}

\newcommand{\orientation}{\glslink{orientation}{\theta}}
\newglossaryentry{orientation}{
	name={\ensuremath{\orientation}},
	description={Orientation in $\cartesianframe$},
	sort=6l,
	type=state
}

\newcommandx{\orientationk}[1][1=\timestep, usedefault]{\glslink{orientationk}{\theta_{#1}}}
\newglossaryentry{orientationk}{
	name={\ensuremath{\orientationk}},
	description={Orientation $\orientation$ at time step $\timestep$},
	sort=6l,
	type=state,
	parent=orientation
}

\newcommand{\curvature}{\glslink{curvature}{\kappa}}
\newglossaryentry{curvature}{
	name={\ensuremath{\curvature}},
	description={Curvature in $\cartesianframe$},
	sort=6l,
	type=state
}

\newcommandx{\curvaturek}[1][1=\timestep, usedefault]{\curvature_{#1}}
\newglossaryentry{curvaturek}{
	name={\ensuremath{\curvaturek}},
	description={Curvature $\curvature$ at time step $\timestep$},
	sort=6l,
	type=state,
	parent=curvature
}

\newcommand{\refpathsym}{\Gamma}

\newcommandx{\refpath}[1][1=\slon, usedefault]{\glslink{refpath}{\refpathsym(#1)}}
\newglossaryentry{refpath}{
	name={\ensuremath{\refpath}},
	description={Reference path},
	sort=7a,
	type=refpath
}

\newcommand{\reforientation}{\glslink{reforientation}{\orientation_\refpathsym}}
\newglossaryentry{reforientation}{
	name={\ensuremath{\reforientation}},
	description={Orientation of reference path},
	sort=7b,
	type=refpath
}

\newcommand{\refcurvature}{\glslink{refcurvature}{\curvature_\refpathsym}}
\newglossaryentry{refcurvature}{
	name={\ensuremath{\refcurvature}},
	description={Curvature of reference path},
	sort=7c,
	type=refpath
}

\newcommand{\minrefcurvature}{\glslink{minrefcurvature}{\minval{\curvature}_\refpathsym}}
\newglossaryentry{minrefcurvature}{
	name={\ensuremath{\minrefcurvature}},
	description={Minimum curvature of reference path},
	sort=7d,
	type=refpath
}

\newcommand{\maxrefcurvature}{\glslink{maxrefcurvature}{\maxval{\curvature}_\refpathsym}}
\newglossaryentry{maxrefcurvature}{
	name={\ensuremath{\maxrefcurvature}},
	description={Maximum curvature of reference path},
	sort=7f,
	type=refpath
}

\newcommand{\idxcentercircle}{i}
\newcommand{\centerpoint}{c}

\newcommand{\wheelbase}{\glslink{wheelbase}{\ell}}
\newglossaryentry{wheelbase}{
	name={\ensuremath{\wheelbase}},
	description={Wheelbase},
	sort=8a,
	type=shape
}

\newcommand{\radius}{\glslink{radius}{r}}
\newglossaryentry{radius}{
	name={\ensuremath{\radius}},
	description={Radius of the circles approximating the vehicle shape},
	sort=8b,
	type=shape
}

\newcommandx{\centercircle}[2][1=\idxcentercircle, 2=\timestep, usedefault]{\glslink{centercircle}{\centerpoint^{(#1)}_{#2}}\glslink{centercirclek}{}}
\newglossaryentry{centercircle}{
	name={\ensuremath{\centercircle[][ ]}},
	description={Center of the $\idxcentercircle$-th circle approximating the vehicle shape},
	sort=8c,
	type=shape
}
\newglossaryentry{centercirclek}{
	name={\ensuremath{\centercircle}},
	description={Center $\centercircle$ at time step $\timestep$},
	sort=8c,
	type=shape,
	parent=centercircle,
}

\newcommandx{\globalcentercirclek}[2][1=\idxcentercircle, 2=\timestep, usedefault]{\glslink{globalcentercirclek}{\globalvar{\centerpoint}^{(#1)}_{#2}}}
\newglossaryentry{globalcentercirclek}{
	name={\ensuremath{\globalcentercirclek}},
	description={Center $\centercircle$ in $\cartesianframe$ at time step $\timestep$},
	sort=8c,
	type=shape,
	parent=centercircle
}

\newcommandx{\globalcentercirclerayk}[2][1=\idxcentercircle, 2=\timestep, usedefault]{\glslink{globalcentercirclerayk}{\globalvar{\hat{\centerpoint}}^{(#1)}_{#2}}}
\newglossaryentry{globalcentercirclerayk}{
	name={\ensuremath{\globalcentercirclerayk}},
	description={Center $\centercircle$ in $\cartesianframe$ at time step $\timestep$},
	sort=8c,
	type=shape,
	parent=centercircle
}

\newcommandx{\localcentercirclek}[2][1=\idxcentercircle, 2=\timestep, usedefault]{\glslink{localcentercirclek}{\localvar{\centerpoint}^{(#1)}_{#2}}}
\newglossaryentry{localcentercirclek}{
	name={\ensuremath{\localcentercirclek}},
	description={Center $\centercircle$ in $\curvframe$ at time step $\timestep$},
	sort=8c,
	type=shape,
	parent=centercircle
}

\newcommand{\cut}{\mathrm{cut}}

\newcommand{\conttime}{\glslink{time}{t}}
\newglossaryentry{time}{
	name={\ensuremath{\conttime}},
	description={Time},
	sort=1 a,
	type=time,
}

\newcommand{\dt}{\glslink{dt}{\Delta \conttime}}
\newglossaryentry{dt}{
	name={\ensuremath{\dt}},
	description={Time increment},
	sort=1 b,
	type=time,
}

\newcommand{\thorizon}{\glslink{thorizon}{h}}
\newglossaryentry{thorizon}{
	name={\ensuremath{\thorizon}},
	description={Final time step},
	sort=1 c,
	type=time
}

\newcommand{\tinit}{\glslink{tinit}{0}}
\newglossaryentry{tinit}{
	name={\ensuremath{\tinit}},
	description={Initial time step},
	sort=1 d,
	type=time
}

\newcommand{\tcutoff}{\glslink{tinit}{\cut}}
\newglossaryentry{tcutoff}{
	name={\ensuremath{\tcutoff}},
	description={Cut-off time step},
	sort=1 h,
	type=time
}

\newcommand{\timestep}{\glslink{timestep_b}{k}}
\newglossaryentry{timestep_a}{
	name=\ensuremath{\timestep},
	description={Discrete time},
	sort=1 f,
	type=time,
}
\newglossaryentry{timestep_b}{
	name={\ensuremath{\square_k}},
	description={Variable $\square$ at time step $\timestep$},
	parent=timestep_a,
	sort=1 g,
	type=time,
}

\newcommand{\lon}{\mathrm{lon}}
\newcommand{\lat}{\mathrm{lat}}
\newcommand{\R}{\mathcal{R}}

\newcommand{\systemmatrix}{\glslink{systemmatrix}{A}}
\newglossaryentry{systemmatrix}{
	name={\ensuremath{\systemmatrix}},
	description={System matrix},
	sort=c e
}

\newcommand{\inputmatrix}{\glslink{inputmatrix}{B}}
\newglossaryentry{inputmatrix}{
	name={\ensuremath{\inputmatrix}},
	description={Input matrix},
	sort=c f
}

\newcommand{\x}{\glslink{x}{x}}
\newglossaryentry{x}{
	name={\ensuremath{\x}},
	description={State},
	sort=c g
}

\newcommand{\systeminput}{\glslink{systeminput}{u}}
\newglossaryentry{systeminput}{
	name={\ensuremath{\systeminput}},
	description={Input},
	sort=c h
}

\newcommand{\X}{\glslink{X}{\mathcal{X}}}
\newglossaryentry{X}{
	name={\ensuremath{\X}},
	description={Admissible set of states},
	sort=c i
}

\newcommand{\U}{\glslink{U}{\mathcal{U}}}
\newglossaryentry{U}{
	name={\ensuremath{\U}},
	description={Admissible set of inputs},
	sort=c j
}

\newcommandx{\initialstatetrajectory}{\glslink{initialstatetrajectory}{\chi^{\mathtt{int}}}}
\newglossaryentry{initialstatetrajectory}{
	name={\ensuremath{\initialstatetrajectory[ ]}},
	description={State trajectory},
	sort=c k
}

\newcommandx{\repairedstatetrajectory}{\glslink{repairedstatetrajectory}{\chi^{\mathtt{rep}}}}
\newglossaryentry{repairedstatetrajectory}{
	name={\ensuremath{\repairedstatetrajectory[ ]}},
	description={State trajectory},
	sort=c k
}

\newcommandx{\initialinputtrajectory}{\glslink{initialinputtrajectory}{\chi^{\mathtt{int}}}}
\newglossaryentry{initialinputtrajectory}{
	name={\ensuremath{\initialinputtrajectory[ ]}},
	description={State trajectory},
	sort=c k
}

\newcommandx{\repairedinputtrajectory}{\glslink{repairedinputtrajectory}{\chi^{\mathtt{rep}}}}
\newglossaryentry{repairedinputtrajectory}{
	name={\ensuremath{\repairedinputtrajectory[ ]}},
	description={State trajectory},
	sort=c k
}

\newcommandx{\statetrajectory}[1][1=i, usedefault]{\glslink{statetrajectory}{X_{#1}}}
\newglossaryentry{statetrajectory}{
	name={\ensuremath{\statetrajectory[ ]}},
	description={State trajectory},
	sort=c k
}

\newcommandx{\inputtrajectory}[1][1=i, usedefault]{\glslink{inputtrajectory}{U_{#1}}}
\newglossaryentry{inputtrajectory}{
	name={\ensuremath{\inputtrajectory[ ]}},
	description={Input trajectory},
	sort=c l
}

\newcommand{\utotal}{\glslink{utotal}{\systeminput}}
\newglossaryentry{utotal}{
	name={\ensuremath{\utotal}},
	description={Input},
	type=trajectoryrepairing,
	sort=2 e
}

\newcommand{\xtotal}{\glslink{xtotal}{\x}}
\newglossaryentry{xtotal}{
	name={\ensuremath{\xtotal}},
	description={State},
	type=trajectoryrepairing,
	sort=2 j
}

\newcommand{\costtotal}{\glslink{costtotal}{J}}
\newglossaryentry{costtotal}{
	name={\ensuremath{\costtotal}},
	description={Cost function},
	type=trajectoryrepairing,
	sort=2 i
}

\newcommand{\robusttotal}{\glslink{robosttotal}{P}}
\newglossaryentry{robosttotal}{
	name={\ensuremath{\robusttotal}},
	description={Robustness function},
	type=trajectoryrepairing,
	sort=2 h
}

\newcommand{\xinit}{\glslink{xinit}{\x_0}}
\newglossaryentry{xinit}{
	name={\ensuremath{\xinit}},
	description={Initial trajectory},
	type=trajectoryrepairing,
	sort=2 j
}

\newcommandx{\xtotalk}[1][1=\timestep, usedefault]{\glslink{xtotalk}{\xtotal_{{#1}}}}
\newglossaryentry{xtotalk}{
	name={\ensuremath{\xtotalk}},
	description={State at time step $\timestep$},
	parent=xtotal,
	type=trajectoryrepairing,
	sort=2 m
}

\newcommandx{\xrepairk}[1][1=\timestep, usedefault]{\glslink{xrepairk}{\xtotal^{\mathtt{rep}}_{{#1}}}}
\newglossaryentry{xrepairk}{
	name={\ensuremath{\xrepairk}},
	description={State at time step $\timestep$ of the repaired trajectory},
	parent=xtotal,
	type=trajectoryrepairing,
	sort=2 m
}

\newcommandx{\xinitk}[1][1=\timestep, usedefault]{\glslink{xinitk}{\xtotal^{\mathtt{int}}_{{#1}}}}
\newglossaryentry{xinitk}{
	name={\ensuremath{\xinitk}},
	description={State at time step $\timestep$ of the initial trajectory},
	parent=xtotal,
	type=trajectoryrepairing,
	sort=2 m
}

\newcommandx{\uinitk}[1][1=\timestep, usedefault]{\glslink{uinitk}{\utotal^{\mathtt{int}}_{{#1}}}}
\newglossaryentry{uinitk}{
	name={\ensuremath{\uinitk}},
	description={Input at time step $\timestep$ of the initial trajectory},
	parent=xtotal,
	type=trajectoryrepairing,
	sort=2 m
}

\newcommandx{\utotalk}[1][1=\timestep, usedefault]{\glslink{utotalk}{\utotal_{{#1}}}}
\newglossaryentry{utotalk}{
	name={\ensuremath{\utotalk}},
	description={Input at time step $\timestep$},
	parent=utotal,
	type=trajectoryrepairing,
	sort=2 n
}

\newcommandx{\urepairk}[1][1=\timestep, usedefault]{\glslink{urepairk}{\utotal^{\mathtt{rep}}_{{#1}}}}
\newglossaryentry{urepairk}{
	name={\ensuremath{\urepairk}},
	description={Input at time step $\timestep$ of the repaired trajectory},
	parent=utotal,
	type=trajectoryrepairing,
	sort=2 m
}

\newcommandx{\Xtotalk}[1][1=\timestep, usedefault]{\glslink{Xtotalk}{\X_{#1}}}
\newglossaryentry{Xtotalk}{
	name={\ensuremath{\Xtotalk}},
	description={Set of admissible states at time step $\timestep$},
	type=trajectoryrepairing,
	sort=2 f
}

\newcommandx{\Utotalk}[1][1=\timestep, usedefault]{\glslink{Utotalk}{\U_{#1}}}
\newglossaryentry{Utotalk}{
	name={\ensuremath{\Utotalk}},
	description={Set of admissible inputs at time step $\timestep$},
	type=trajectoryrepairing,
	sort=2 g
}

\newcommand{\reppredicate}{\glslink{reppredicate}{\hat{\psi}}}

\newglossaryentry{reppredicate}{
	name={\ensuremath{\reppredicate}},
	description={State},
	type=trajectoryrepairing,
	sort=2 j
}

\newcommand{\Alon}{\glslink{Alon}{\systemmatrix_{\lon}}}
\newglossaryentry{Alon}{
	name={\ensuremath{\Alon}},
	description={System matrix},
	type=longitudinalplanning,
	sort=2 a
}
\newcommandx{\Alonk}[1][1=\timestep, usedefault]{\glslink{Alonk}{\systemmatrix_{\lon, #1}}}
\newglossaryentry{Alonk}{
	name={\ensuremath{\Alonk}},
	description={System matrix at time step $\timestep$},
	parent=Alon,
	type=longitudinalplanning,
	sort=2 a
}

\newcommandx{\Blon}{\glslink{Blon}{\inputmatrix_{\lon}}}
\newglossaryentry{Blon}{
	name={\ensuremath{\Blon}},
	description={Input matrix trajectory planning},
	type=longitudinalplanning,
	sort=2 b
}
\newcommandx{\Blonk}[1][1=\timestep, usedefault]{\glslink{Blonk}{\inputmatrix_{\lon, #1}}}
\newglossaryentry{Blonk}{
	name={\ensuremath{\Blonk}},
	description={Input matrix at time step $\timestep$},
	parent=Blon,
	type=longitudinalplanning,
	sort=2 b
}

\newcommand{\xlon}{\glslink{xlon}{\x_\lon}}
\newglossaryentry{xlon}{
	name={\ensuremath{\xlon}},
	description={State},
	type=longitudinalplanning,
	sort=2 c
}

\newcommandx{\xlonk}[1][1=\timestep, usedefault]{\glslink{xlonk}{\x_{{\lon, #1}}}}
\newglossaryentry{xlonk}{
	name={\ensuremath{\xlonk}},
	description={State at time step $\timestep$},
	parent=xlon,
	type=longitudinalplanning,
	sort=2 c
}

\newcommand{\ulon}{\glslink{ulon}{\systeminput_\lon}}
\newglossaryentry{ulon}{
	name={\ensuremath{\ulon}},
	description={Input},
	type=longitudinalplanning,
	sort=2 e
}

\newcommandx{\ulonk}[1][1=\timestep, usedefault]{\glslink{ulonk}{\systeminput_{\lon,#1}}}
\newglossaryentry{ulonk}{
	name={\ensuremath{\ulonk}},
	description={Input at time step $\timestep$},
	parent=ulon,
	type=longitudinalplanning,
	sort=2 e
}

\newcommandx{\Xlonk}[1][1=\timestep, usedefault]{\glslink{Xlonk}{\X_{\lon, #1}}}
\newglossaryentry{Xlonk}{
	name={\ensuremath{\Xlonk}},
	description={Set of admissible states at time step $\timestep$},
	type=longitudinalplanning,
	sort=2 f
}

\newcommandx{\Ulonk}[1][1=\timestep, usedefault]{\glslink{Ulonk}{\U_{\lon, #1}}}
\newglossaryentry{Ulonk}{
	name={\ensuremath{\Ulonk}},
	description={Set of admissible inputs at time step $\timestep$},
	type=longitudinalplanning,
	sort=2 g
}

\newcommandx{\statetrajectorylon}{\glslink{statetrajectorylon}{\statetrajectory[\lon]}}
\newglossaryentry{statetrajectorylon}{
	name={\ensuremath{\statetrajectorylon}},
	description={Longitudinal state trajectory},
	type=longitudinalplanning,
	sort=2 h
}

\newcommand{\costlon}{\glslink{costlon}{J_\lon}}
\newglossaryentry{costlon}{
	name={\ensuremath{\costlon}},
	description={Cost function},
	type=longitudinalplanning,
	sort=2 i
}

\newcommand{\Alat}{\glslink{Alat}{\systemmatrix_{\lat}}}
\newglossaryentry{Alat}{
	name={\ensuremath{\Alat}},
	description={System matrix},
	sort=3 a,
	type=lateralplanning,
}
\newcommandx{\Alatk}[1][1=\timestep, usedefault]{\glslink{Alatk}{\systemmatrix_{\lat, #1}}}
\newglossaryentry{Alatk}{
	name={\ensuremath{\Alatk}},
	description={System matrix at time step $\timestep$},
	sort=3 b,
	parent=Alat,
	type=lateralplanning,
}

\newcommandx{\Blat}{\glslink{Blat}{\inputmatrix_{\lat}}}
\newglossaryentry{Blat}{
	name={\ensuremath{\Blat}},
	description={Input matrix},
	sort=3 c,
	type=lateralplanning,
}
\newcommandx{\Blatk}[1][1=\timestep, usedefault]{\glslink{Blatk}{\inputmatrix_{\lat, #1}}}
\newglossaryentry{Blatk}{
	name={\ensuremath{\Blatk}},
	description={Input matrix at time step $\timestep$},
	sort=3 d,
	parent=Blat,
	type=lateralplanning,
}

\newcommand{\xlat}{\glslink{xlat}{\x_\lat}}
\newglossaryentry{xlat}{
	name={\ensuremath{\xlat}},
	description={State},
	sort=3 f,
	type=lateralplanning,
}

\newcommandx{\xlatk}[1][1=\timestep, usedefault]{\glslink{xlatk}{\x_{{\lat, #1}}}}
\newglossaryentry{xlatk}{
	name={\ensuremath{\xlatk}},
	description={State at time step $\timestep$},
	sort=3 g,
	parent=xlat,
	type=lateralplanning,
}

\newcommand{\ulat}{\glslink{ulat}{\systeminput_\lat}}
\newglossaryentry{ulat}{
	name={\ensuremath{\ulat}},
	description={Input},
	sort=3 h,
	type=lateralplanning,
}

\newcommandx{\ulatk}[1][1=\timestep, usedefault]{\glslink{ulatk}{\systeminput_{\lat,#1}}}
\newglossaryentry{ulatk}{
	name={\ensuremath{\ulatk}},
	description={Input at time step $\timestep$},
	sort=3 i,
	type=lateralplanning,
	parent=ulat
}

\newcommandx{\Xlatk}[1][1=\timestep, usedefault]{\glslink{Xlatk}{\X_{\lat, #1}}}
\newglossaryentry{Xlatk}{
	name={\ensuremath{\Xlatk}},
	description={Set of admissible states at time step $\timestep$},
	type=lateralplanning,
	sort=3 j
}

\newcommandx{\Ulatk}[1][1=\timestep, usedefault]{\glslink{Ulatk}{\U_{\lat, #1}}}
\newglossaryentry{Ulatk}{
	name={\ensuremath{\Ulatk}},
	description={Set of admissible inputs at time step $\timestep$},
	type=lateralplanning,
	sort=3 k
}

\newcommandx{\statetrajectorylat}{\glslink{statetrajectorylat}{\statetrajectory[\lat]}}
\newglossaryentry{statetrajectorylat}{
	name={\ensuremath{\statetrajectorylat}},
	description={Lateral state trajectory},
	type=lateralplanning,
	sort=3 l
}

\newcommand{\costlat}{\glslink{costlat}{J_\lat}}
\newglossaryentry{costlat}{
	name={\ensuremath{\costlat}},
	description={Cost function},
	type=lateralplanning,
	sort=3 m
}

\newcommandx{\obstaclesk}[1][1=\globalvar, usedefault]{\glslink{obstaclesk}{#1{\mathcal{O}}_{\timestep}}}
\newglossaryentry{obstaclesk}{
	name={\ensuremath{\obstaclesk}},
	description={Occupancy sets of all obstacles in $\cartesianframe$ at time step $\timestep$},
	type=obs,
	sort=9a
}

\newcommandx{\obstaclescirclek}[1][1=\globalvar, usedefault]{\glslink{obstaclescirclek}{#1{\mathcal{O}}^{\basiccircleshape}_{\timestep}}}
\newglossaryentry{obstaclescirclek}{
	name={\ensuremath{\obstaclescirclek}},
	description={Occupancy sets $\obstaclesk$ dilated with circle $\basiccircleshape$},
	type=obs,
	sort=9c
}

\newcommand{\Areach}{\glslink{Areach}{\systemmatrix_{\R}}}
\newglossaryentry{Areach}{
	name={\ensuremath{\Areach}},
	description={System matrix},
	sort=10a,
	type=reach,
}

\newcommand{\Breach}{\glslink{Breach}{\inputmatrix_{\R}}}
\newglossaryentry{Breach}{
	name={\ensuremath{\Breach}},
	description={Input matrix},
	sort=10b,
	type=reach,
}

\newcommand{\xreach}{\glslink{xreach}{\x_{\R}}}
\newglossaryentry{xreach}{
	name={\ensuremath{\xreach}},
	description={State},
	sort=10c,
	type=reach,
}

\newcommandx{\xreachk}[1][1=\timestep, usedefault]{\glslink{xreachk}{\x_{\R, #1}}}
\newglossaryentry{xreachk}{
	name={\ensuremath{\xreachk}},
	description={State $\xreach$ at time step $\timestep$},
	sort=10c,
	type=reach,
	parent=xreach
}

\newcommand{\ureach}{\glslink{ureach}{\glslink{ureach}{\systeminput_{\R}}}}
\newglossaryentry{ureach}{
	name={\ensuremath{\ureach}},
	description={Input},
	sort=10d,
	type=reach,
}

\newcommandx{\ureachk}[1][1=\timestep, usedefault]{\glslink{ureachk}{\systeminput_{\R, #1}}}
\newglossaryentry{ureachk}{
	name={\ensuremath{\ureachk}},
	description={Input $\ureach$ at time step $\timestep$},
	sort=10d,
	type=reach,
	parent=ureach
}

\newcommandx{\Xreachk}[1][1=\timestep, usedefault]{\glslink{Xreachk}{\X_{\R, #1}}}
\newglossaryentry{Xreachk}{
	name={\ensuremath{\Xreachk}},
	description={Set of admissible states at time step $\timestep$},
	type=reach,
	sort=10e
}

\newcommandx{\Ureachk}[1][1=\timestep, usedefault]{\glslink{Ureachk}{\U_{\R, #1}}}
\newglossaryentry{Ureachk}{
	name={\ensuremath{\Ureachk}},
	description={Set of admissible inputs at time step $\timestep$},
	type=reach,
	sort=10f
}

\newcommandx{\forbiddenset}[1][1=\timestep, usedefault]{\glslink{forbiddenset}{\mathcal{F}_{#1}}}
\newglossaryentry{forbiddenset}{
	name={\ensuremath{\forbiddenset}},
	description={Set of forbidden states at time step $\timestep$},
	type=reach,
	sort=10g
}

\newcommandx{\egooccupancy}[2][1={(\xreachk)}, 2=\globalvar, usedefault]{\glslink{egooccupancy}{#2{\mathcal{Q}}#1}}
\newglossaryentry{egooccupancy}{
	name={\ensuremath{\egooccupancy}},
	description={Occupancy of ego vehicle in $\cartesianframe$ at state $\xreachk$},
	type=reach,
	sort=10h
}

\newcommand{\exact}{\mathtt{e}}
\newcommand{\exactval}[1]{#1^{\exact}}

\newcommandx{\exactreach}[1][1=\timestep, usedefault]{\glslink{exactreach}{\exactval{\R}_{#1}}}
\newglossaryentry{exactreach}{
	name={\ensuremath{\exactreach}},
	description={Exact reachable set at time step $\timestep$},
	type=reach,
	sort=10i
}

\newcommandx{\reachk}[1][1=\timestep, usedefault]{\glslink{reachk}{\R_{#1}}}
\newglossaryentry{reachk}{
	name={\ensuremath{\reachk}},
	description={Approx. reachable set at time step $\timestep$},
	type=reach,
	sort=10j
}

\newcommand{\D}{\mathcal{D}}
\newcommandx{\exactdrivablearea}[1][1=\timestep, usedefault]{\glslink{exactdrivablearea}{\exactval{\D}_{#1}}}
\newglossaryentry{exactdrivablearea}{
	name={\ensuremath{\exactdrivablearea}},
	description={Exact drivable area at time step $\timestep$},
	type=reach,
	sort=10k
}

\newcommandx{\drivableareak}[1][1=\timestep, usedefault]{\glslink{drivableareak}{\D_{#1}}}
\newglossaryentry{drivableareak}{
	name={\ensuremath{\drivableareak}},
	description={Approx. drivable area at time step $\timestep$},
	type=reach,
	sort=10l
}

\newcommand{\graph}{\mathcal{G}}

\newcommand{\graphreachability}{\glslink{graphreachability}\graph_{\R}}
\newglossaryentry{graphreachability}{
	name={\ensuremath{\graphreachability}},
	description={Reachability graph},
	type=reach,
	sort=10m
}

\newcommand{\idxreach}{i}
\newcommand{\baseset}{\R}

\newcommandx{\sBik}[2][1=(\idxreach), 2=\timestep, usedefault]{\glslink{sBik}{\baseset^{#1}_{#2}}}
\newglossaryentry{sBik}{
	name={\ensuremath{\sBik}},
	description={$\idxreach$-th base set at time step $\timestep$},
	type=reach,
	sort=10n
}

\newcommand{\polytope}{\mathcal{P}}
\newcommandx{\sPikl}[3][1=(\idxreach), 2=\timestep, usedefault]{\glslink{sPikllondir}{\polytope^{#1}_{#3, #2}} \glslink{sPikllatdir}{}}
\newglossaryentry{sPikllondir}{
	name={\ensuremath{\sPikl{\curvframelondir}}},
	description={$\idxreach$-th polytope in the $(\slon, \vlon)$ plane at time step $\timestep$},
	type=reach,
	sort=10o
}
\newglossaryentry{sPikllatdir}{
	name={\ensuremath{\sPikl{\curvframelatdir}}},
	description={$\idxreach$-th polytope in the $(\slat, \vlat)$ plane at time step $\timestep$},
	type=reach,
	sort=10p
}

\newcommand{\aabb}{\D}
\newcommandx{\sAik}[2][1=(\idxreach), 2=\timestep, usedefault]{\glslink{sAik}{\aabb^{#1}_{#2}}}
\newglossaryentry{sAik}{
	name={\ensuremath{\sAik}},
	description={Projection of $\sBik$ onto position domain},
	type=reach,
	sort=10q
}

\newcommand{\prop}{\mathtt{prop}}
\newcommandx{\propreachk}[1][1=\timestep, usedefault]{\glslink{propreachk}{\R^{\prop}_{#1}}}
\newglossaryentry{propreachk}{
	name={\ensuremath{\propreachk}},
	description={Reachable set after propagation step},
	type=reach,
	parent=reachk
}

\newcommandx{\propdrivableareak}[1][1=\timestep, usedefault]{\glslink{propdrivableareak}{\D^{\prop}_{#1}}}
\newglossaryentry{propdrivableareak}{
	name={\ensuremath{\propdrivableareak}},
	description={Drivable area after propagation step},
	type=reach,
	parent=drivableareak
}

\newcommandx{\sBPik}[2][1=(\idxreach), 2=\timestep, usedefault]{\glslink{sBPik}{\baseset^{\prop#1}_{#2}}}
\newglossaryentry{sBPik}{
	name={\ensuremath{\sBPik}},
	description={Propagated base set at time step $\timestep$},
	type=reach,
	parent=sBik
}

\newcommandx{\sAPik}[2][1=(\idxreach), 2=\timestep, usedefault]{\glslink{sAPik}{\aabb^{\prop#1}_{#2}}}
\newglossaryentry{sAPik}{
	name={\ensuremath{\sAPik}},
	description={Projection of $\sBPik$ onto position domain},
	type=reach,
	parent=sAik
}

\newcommandx{\sPPikl}[3][1=(\idxreach), 2=\timestep, usedefault]{\glslink{sPPikllondir}{\polytope^{\prop#1}_{#3, #2}} \glslink{sPPikllatdir}{}}
\newglossaryentry{sPPikllondir}{
	name={\ensuremath{\sPPikl{\curvframelondir}}},
	description={$\idxreach$-th propagated polytope},
	type=reach,
	parent=sPikllondir
}
\newglossaryentry{sPPikllatdir}{
	name={\ensuremath{\sPPikl{\curvframelatdir}}},
	description={$\idxreach$-th propagated polytope},
	type=reach,
	parent=sPikllatdir
}

\newcommand{\repartitioned}{\mathtt{rprt}}
\newcommandx{\repartitioneddrivableareak}[1][1=\timestep, usedefault]{\glslink{repartitioneddrivableareak}{\D^{\repartitioned}_{#1}}}
\newglossaryentry{repartitioneddrivableareak}{
	name={\ensuremath{\repartitioneddrivableareak}},
	description={Drivable area after re-partitioning step},
	type=reach,
	parent=drivableareak,
	sort=10l
}

\newcommandx{\sARik}[2][1=(\idxreach), 2=\timestep, usedefault]{\glslink{sARik}{\aabb^{\repartitioned#1}_{#2}}}
\newglossaryentry{sARik}{
	name={\ensuremath{\sARik[(q)]}},
	description={Part of drivable area after re-partitioning step},
	type=reach,
	parent=sAik,
}

\newcommandx{\sPPhatikl}[3][1=(\idxreach), 2=\timestep, usedefault]{\hat{\polytope}^{#1}_{#3, #2}}

\newcommand{\enlargement}{\glslink{enlargement}{\mathcal{A}^\epsilon}}
\newglossaryentry{enlargement}{
	name={\ensuremath{\enlargement}},
	description={Enlargement to consider shape of ego vehicle},
	type=reach,
	sort=10r
}

\newcommand{\idxconnectedset}{n}
\newcommand{\connected}{\mathcal{C}}

\newcommand{\connectedset}{\glslink{connectedset}{\mathcal{C}}}
\newglossaryentry{connectedset}{
	name={\ensuremath{\connectedset}},
	description={Connected set in the position domain},
	type=drivingcorridor,
	sort=11a
}

\newcommandx{\connectedsetk}[1][1=\timestep, usedefault]{\glslink{connectedsetk}{\connectedset_{#1}}}
\newglossaryentry{connectedsetk}{
	name={\ensuremath{\connectedsetk}},
	description={Connected set at time step $\timestep$},
	type=drivingcorridor,
	parent=connectedset
}

\newcommandx{\connectedsetkn}[2][1=\timestep, 2=\idxconnectedset, usedefault]{\glslink{connectedsetkn}{\connectedset^{(#2)}_{#1}}}
\newglossaryentry{connectedsetkn}{
	name={\ensuremath{\connectedsetkn[][i]}},
	description={$i$-th connected set at time step $\timestep$},
	type=drivingcorridor,
	parent=connectedset
}

\newcommandx{\graphconnectedcomponents}[1][1= , usedefault]{\glslink{graphconnectedcomponents}{\graph_{\connected#1}}}
\newglossaryentry{graphconnectedcomponents}{
	name={\ensuremath{\graphconnectedcomponents}},
	description={Graph storing connected sets},
	type=drivingcorridor,
	sort=11c
}

\newcommand{\node}{\glslink{node}{\mathtt{n}}}
\newglossaryentry{node}{
	name={\ensuremath{\node}},
	description={Node in $\graphconnectedcomponents$},
	type=drivingcorridor,
	parent=graphconnectedcomponents
}

\newcommand{\dc}{C}

\newcommand{\dclon}{\glslink{dclon}{\dc_{\lon}}}
\newglossaryentry{dclon}{
	name={\ensuremath{\dclon}},
	description={Longitudinal driving corridor},
	type=drivingcorridor,
	sort=11e
}

\newcommandx{\dclonk}[1][1=\timestep, usedefault]{\glslink{dclonk}\dc_{\lon, #1}}
\newglossaryentry{dclonk}{
	name={\ensuremath{\dclonk}},
	description={Longitudinal driving corridor at time step $\timestep$},
	type=drivingcorridor,
	parent=dclon
}

\newcommand{\dclat}{\glslink{dclat}{\dc_{\lat}}}
\newglossaryentry{dclat}{
	name={\ensuremath{\dclat}},
	description={Lateral driving corridor},
	type=drivingcorridor,
	sort=11f
}

\newcommandx{\dclatk}[1][1=\timestep, usedefault]{\glslink{dclatk}\dc_{\lat, #1}}
\newglossaryentry{dclatk}{
	name={\ensuremath{\dclatk}},
	description={Lateral driving corridor at time step $\timestep$},
	type=drivingcorridor,
	parent=dclat
}

\newcommandx{\parentset}[1][1=k-1, usedefault]{\mathcal{D}^{\mathrm{parents}}_{#1}}

\newcommand{\distance}{d}
\newcommandx{\latdiscircle}[2][1=\idxcentercircle, 2=\timestep, usedefault]{\glslink{latdiscircle}{\distance^{(#1)}_{#2}}}
\newglossaryentry{latdiscircle}{
	name={\ensuremath{\latdiscircle}},
	description={Lateral distance of $\centercircle$ from $\refpath$ at time step $\timestep$},
	type=constraints,
	sort=12a
}

\newcommandx{\dmink}[2][1=\idxcentercircle, 2=\timestep, usedefault]{\glslink{dmink}{\minval{\distance}^{(#1)}_{#2}}}
\newglossaryentry{dmink}{
	name={\ensuremath{\dmink}},
	description={Minimum admissible value of $\latdiscircle$},
	type=constraints,
	parent=latdiscircle
}

\newcommandx{\dmaxk}[2][1=\idxcentercircle, 2=\timestep, usedefault]{\glslink{dmaxk}{\maxval{\distance}^{(#1)}_{#2}}}
\newglossaryentry{dmaxk}{
	name={\ensuremath{\dmaxk}},
	description={Maximum admissible value of $\latdiscircle$},
	type=constraints,
	parent=latdiscircle
}

\newcommandx{\straightlineik}[3][1=\idxcentercircle, 2=\timestep, usedefault]{\glslink{straightlineik}{g^{(#1)}_{#2}#3}}
\newglossaryentry{straightlineik}{
	name={\ensuremath{\straightlineik{}}},
	description={Straight line perpendicular to $\refpath$ going through $\centercircle$},
	type=constraints,
	sort=12b
}

\newcommandx{\intersectingposik}[2][1=\idxcentercircle, 2=\timestep, usedefault]{\glslink{intersectingposik}{\mathcal{Y}^{(#1)}_{#2}}}
\newglossaryentry{intersectingposik}{
	name={\ensuremath{\intersectingposik}},
	description={Positions in $\dclon$ intersecting with $\straightlineik{}$},
	type=constraints,
	sort=12c
}

\newcommand{\idxinterval}{q}
\newcommandx{\intervaliqk}[3][1=\idxcentercircle, 2=\timestep, 3=\idxinterval, usedefault]{\glslink{intervaliqk}{\mathcal{I}^{(#1)}_{#2,#3}}}
\newglossaryentry{intervaliqk}{
	name={\ensuremath{\intervaliqk}},
	description={$\idxinterval$-th interval of positions in $\dclon$ intersecting $\straightlineik{}$},
	type=constraints,
	sort=12d
}

\newcommandx{\validintervals}[1][1=\idxcentercircle,usedefault]{\mathcal{I}^{(#1)}_{\mathtt{v}}}

\newcommand{\placeholdersys}{\glslink{placeholdersys}{\mathrm{sys}}}
\newglossaryentry{placeholdersys}{
	name={\ensuremath{\placeholdersys}},
	description={Placeholder for a variable},
	sort=a
}

\newcommandx{\proj}[2][1=\lozenge, usedefault]{\glslink{proj}{\mathrm{proj}_{#1}\!(#2 )}}
\newglossaryentry{proj}{
	name={\ensuremath{\proj[\lozenge]{\x}}},
	description={Projection operator that maps the state $\x$ onto its elements $\lozenge$},
	sort=z
}

\newcommandx{\overlap}[1][1=\sAik, usedefault]{\glslink{overlap}{\mathrm{overlap}(#1 )}}
\newglossaryentry{overlap}{
	name={\ensuremath{\overlap}},
	description={Returns all indices $\idxreach$ of the propagated sets $\sBPik$ that overlap with $\sAik$},
	sort=z
}

\newcommand{\convexhull}{\glslink{convexhull}{\mathrm{convexhull}}}
\newglossaryentry{convexhull}{
	name={\ensuremath{\convexhull}},
	description={Computes the convex hull},
	sort=z
}

\newcommand{\signal}{S}

\newcommandx{\signalk}[1][1=\timestep, usedefault]{\glslink{signalk}{\signal_{{#1}}}}
\newglossaryentry{signalk}{
	name={\ensuremath{\signalk}},
	description={Signal at time step $\timestep$},
	type=signaltemporallogic,
	sort=14 a
}

\newcommandx{\sminlonARik}[2][1=q, 2=\timestep, usedefault]{\minval{\pos}^{\repartitioned(#1)}_{\curvframelondir, #2}}
\newcommandx{\smaxlonARik}[2][1=q, 2=\timestep, usedefault]{\maxval{\pos}^{\repartitioned(#1)}_{\curvframelondir, #2}}
\newcommandx{\sminlatARik}[2][1=q, 2=\timestep, usedefault]{\minval{\pos}^{\repartitioned(#1)}_{\curvframelatdir, #2}}
\newcommandx{\smaxlatARik}[2][1=q, 2=\timestep, usedefault]{\maxval{\pos}^{\repartitioned(#1)}_{\curvframelatdir, #2}}

\newcommand{\approxsymb}{\mathtt{C}}

\newcommandx{\refcurvaturecirclei}[1][1=i, usedefault]{\curvature_{\refpathsym,#1}^{\approxsymb}}

\usepackage{hyperref}

\setquotestyle{english}
\title{\LARGE \bf
	 Automatic Traffic Scenario Conversion from\\ OpenSCENARIO to CommonRoad
 }

\author{Yuanfei Lin, Michael Ratzel, and Matthias Althoff
	\thanks{The authors are with the School of Computation, Information and Technology, Technical University of Munich, 85748 Garching, Germany.} 
	\thanks{\tt \{yuanfei.lin, michael.ratzel, althoff\}@tum.de}
}

\begin{document}

	\maketitle
	\thispagestyle{empty}
	\pagestyle{empty}

	\begin{abstract}
		Scenarios are a crucial element for developing, testing, and verifying autonomous driving systems. However, open-source scenarios are often formulated using different terminologies. This limits their usage across different applications as many scenario representation formats are not directly compatible with each other. To address this problem, we present the first open-source converter from the OpenSCENARIO format to the CommonRoad format, which are two of the most popular scenario formats used in autonomous driving. Our converter employs a simulation tool to execute the dynamic elements defined by OpenSCENARIO.  The converter is available at \href{https://commonroad.in.tum.de/}{\mbox{commonroad.in.tum.de}} and we demonstrate its usefulness by converting publicly available scenarios in the OpenSCENARIO format and evaluating them using CommonRoad tools.
	\end{abstract}
	
	
	\section{Introduction}\label{sec:introduction}
Scenarios play a significant role in the development, testing, and validation of autonomous driving systems \cite{menzel2018scenarios}. 
However, there is a shortage of both open-source and commonly-used scenarios. Various representation formats for scenarios are supported by different applications, depending on their specific purposes. 
For example,  OpenSCENARIO\footnote{\scriptsize\url{https://www.asam.net/standards/detail/openscenario/}} employs a logical scenario description that consists of a parameterized set of variables. Instead, CommonRoad \cite{Althoff2017a} describes concrete scenarios that are instances of a logical scenario with fixed parameters.
For this reason, a converter between different formats is desired to promote the exchange and usability of scenarios.  In this work, we present the first openly accessible converter from OpenSCENARIO to CommonRoad, two widely-used formats in the field of autonomous driving \cite{riedmaier2020survey}.
\subsection{Related Work}\label{sec:li_ov}
Next, we review different scenario formats and the works that use them, followed by discussing their capabilities. 
\paragraph {\it CommonRoad}CommonRoad scenarios are represented as XML files containing a detailed description of the road network, traffic participant movements, and vehicle planning problems. To facilitate benchmarking of motion planning on roads, CommonRoad provides a range of vehicle models and cost functions. To enable the use of more diverse and realistic scenarios, CommonRoad provides dataset converters\footnote{\scriptsize\url{https://commonroad.in.tum.de/tools/dataset-converters}} to convert real-world data from various sources, such as drones \cite{highd, interactiondataset, bock2020ind,krajewski2020round, moers2022exid, xu2022drone}, onboard sensors \cite{nuplan},
and infrastructure  \cite{gressenbuch2022mona}, into a unified representation. One can also create handcrafted or generate safety-critical traffic scenarios based on that data \cite{maierhofer2021commonroad, Klischat2019b, klischat2020scenario}. In addition, CommonRoad can be coupled with other simulation software platforms such as SUMO \cite{klischat2019coupling} and Apollo \cite{wang2020coupling}  to test motion planning algorithms in interactive driving environments.
The suite of open-source tools provided by CommonRoad is extensive and robust, featuring a drivability checker \cite{PekIV20}, a set-based predictor \cite{KoschiSPOT}, a reachability analyzer \cite{liu2022commonroad}, and a criticality estimator~\cite{YuanfeiLin2023CriMe}. These tools are designed to be effective and user-friendly in evaluating scenarios, making them a convenient option for various applications. For instance, safe, ethical, and robust motion planning algorithms are benchmarked using CommonRoad scenarios in \cite{christian2020nature, nyberg2021risk, li2022motion, geisslinger2023ethical}. The authors in \cite{zanardi2021posetal, xiong2022noncooperative, zanardi2022factorization} utilize CommonRoad tools to demonstrate the game-theoretic aspects of autonomous vehicles.
Furthermore, CommonRoad paves the way to use advanced algorithms to facilitate motion planning, such as reinforcement learning~\cite{commonroad-rl, khaitan2022state} and  geometric deep learning~\cite{meyer2023geometric}. 
\paragraph {\it ASAM OpenX}
ASAM\footnote{\scriptsize\url{https://www.asam.net}} is a standardization organization that defines open file formats (aka OpenX) for autonomous driving and traffic simulation.
OpenSCENARIO specifies the dynamic aspects of the environment, while OpenDRIVE\footnote{\scriptsize\url{https://www.asam.net/standards/detail/opendrive/}} and OpenCRG\footnote{\scriptsize\url{https://www.asam.net/standards/detail/opencrg/}} define the static elements such as road networks. Additionally, the Open Simulation Interface\footnote{\scriptsize\url{https://www.asam.net/standards/detail/osi/}} (OSI) is regarded as a standardized interface that provides easy and straightforward compatibility between different simulation platforms.
There exist several tools publicly available for generating OpenSCENARIO and OpenDRIVE files, such as Open Scenario Editor\footnote{\scriptsize\url{https://github.com/ebadi/OpenScenarioEditor}}, scenariogeneration\footnote{\scriptsize\url{https://github.com/pyoscx/scenariogeneration}}, and MATLAB RoadRunner\footnote{\scriptsize\url{https://mathworks.com/products/roadrunner.html}}. For testing and validating autonomous vehicles, OpenSCENARIO can be utilized in simulation platforms like openPASS \cite{dobberstein2017eclipse}, esmini\footnote{\scriptsize\url{https://github.com/esmini/esmini}}, and CARLA \cite{dosovitskiy2017carla} to execute complex traffic scenarios \cite{tenbrock2021conscend, karunakaran2022parameterisation, chen2022generating, salles2022modular}.
\paragraph {\it Other Scenario Formats}
GeoScenario \cite{queiroz2019geoscenario} is a domain-specific language akin to OpenSCENARIO, designed for constructing test scenarios. Based on perception systems, the authors in \cite{fremont2019scenic} develop the tool Scenic that defines scenarios as a distribution over configurations of obstacles. Several additional scenario formats are developed to cater to diverse use cases. Examples of such formats include SceML \cite{schutt2020sceml}, SDL \cite{zhang2020scenario}, ADSML \cite{du2021towards}, Paracosm \cite{majumdar2021paracosm}, and MetaScenario~\cite{chang2022metascenario}, among others.
\subsection{Contributions}\label{sec:contri}
In our previous work \cite{althoff2018automatic}, we developed a map converter from the OpenDRIVE format to lanelets\cite{Bender2014}. Lanelets are used by CommonRoad to describe the road geometry. We substantially extend this conversion by encoding the dynamic elements of a scenario specified by OpenSCENARIO. Our converter is expected to be valuable to academic groups and industry professionals alike, given the vast number of openly accessible scenarios available in the CommonRoad and OpenSCENARIO formats.

The paper is structured as follows: Sec.~\ref{sec:conversion} provides a detailed explanation of the scenario conversion from OpenSCENARIO to CommonRoad, further elucidated with numerical examples in Sec.~\ref{sec:exp}. The conclusion is in Sec.~\ref{sec:con}.
{
	\renewcommand{\algorithmicrequire}{\textbf{Input:}}
	\renewcommand{\algorithmicensure}{\textbf{Output:}}
	
	\begin{algorithm*}[h!]\small
		\setstretch{1.0}
		\begin{algorithmic}[1]
			\Require Scenario in the {OpenSCENARIO} format (denoted as {OpenSCENARIO}) 
			\Ensure Scenario in the {CommonRoad} format
			\State lanelet\_network: \texttt{LaneletNetwork} $\gets$ \Call{convertOpendriveToLanelets}{{OpenSCENARIO}.{OpenDRIVE}} \Comment{ See \cite[Alg. 1]{althoff2018automatic}}\label{alg:openDrive}
			\State obstacles: \texttt{List}[\texttt{Obstacle}] $\gets$ simulator.\Call{simulate}{OpenSCENARIO}\label{alg:sim}
			\State scenario: \texttt{Scenario} $\gets$ \Call{buildScenario}{obstacles, lanelet\_network}\label{alg:build_scenario}
			\State ego\_vehicle: \texttt{Obstacle} $\gets$ \Call{findEgoVehicle}{obstacles}\label{alg:ego}
			\State scenario.planning\_problem: \texttt{PlanningProblem} $\gets$ \Call{buildPlanningProblem}{ego\_vehicle} \label{alg:plan} \Comment{initial state and goal}
			\State \Return \Call{writeToXMLFile}{scenario}\label{alg:write}
		\end{algorithmic}
		\caption{\small\textsc{OpenSCENARIOToCommonRoad}\Comment{See Sec.~\ref{subsec:imple}}}
		\label{alg:main}
\end{algorithm*}}
	\section{Conversion from OpenSCENARIO to CommonRoad} \label{sec:conversion}
Both OpenSCENARIO and CommonRoad offer freely available scenarios that can be easily customized and adapted as required. These scenarios are structured hierarchically but with different terminologies. This section concisely presents the conversion of logical scenarios from the OpenSCENARIO format to concrete CommonRoad scenarios. This is presented after a brief introduction of both formats, including an outline of their differences, followed by a detailed description of the conversion process.
\subsection{OpenSCENARIO Format}\label{subsec:openscenario}

\begin{figure}[b!]
	\centering 
	\vspace{-4mm}
	\def\svgwidth{0.95\columnwidth}\scriptsize
	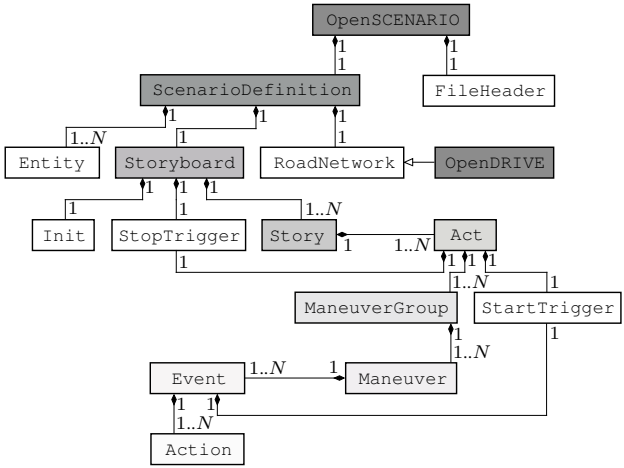
	\caption{UML class diagram of the OpenSCENARIO format.  For brevity, we omit the nonessential classes such as those related to parameters.
	}
	\label{fig:openscenario}
\end{figure}
The architecture of the OpenSCENARIO v1.2.0 format is presented in Fig.~\ref{fig:openscenario} as a unified modeling language (UML) class diagram. The header information for the scenario is contained in the module \texttt{FileHeader}. The class \texttt{ScenarioDefinition} groups the road network (class \texttt{RoadNetwork}), the configuration of obstacles (class \texttt{Entity}), and the container for the dynamic content (class \texttt{StoryBoard}).
The \texttt{StoryBoard} is a core module of OpenSCENARIO as it specifies the temporal sequence of traffic situations and their triggers (class \texttt{Init}, \texttt{StartTrigger}, and \texttt{StopTrigger}) hierarchically into \texttt{Story}, \texttt{Act}, \texttt{Maneuver}, \texttt{Event}, and \texttt{Action}. 

\begin{figure}[t!]	
	\captionsetup[subfigure]{aboveskip=+1mm,belowskip=+1mm}
	\centering
	\begin{subfigure}[b]{\linewidth}
		\centering
		\scriptsize
		\def\svgwidth{\columnwidth}
	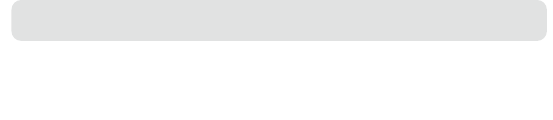
		\caption{{\footnotesize OpenSCENARIO.}}\label{fig:osc_ex}
	\end{subfigure}
	\begin{subfigure}[b]{\linewidth}
		\centering
		\scriptsize
		\def\svgwidth{1\columnwidth}
\begingroup%
  \makeatletter%
  \providecommand\color[2][]{%
    \errmessage{(Inkscape) Color is used for the text in Inkscape, but the package 'color.sty' is not loaded}%
    \renewcommand\color[2][]{}%
  }%
  \providecommand\transparent[1]{%
    \errmessage{(Inkscape) Transparency is used (non-zero) for the text in Inkscape, but the package 'transparent.sty' is not loaded}%
    \renewcommand\transparent[1]{}%
  }%
  \providecommand\rotatebox[2]{#2}%
  \newcommand*\fsize{\dimexpr\f@size pt\relax}%
  \newcommand*\lineheight[1]{\fontsize{\fsize}{#1\fsize}\selectfont}%
  \ifx\svgwidth\undefined%
    \setlength{\unitlength}{265.59289137bp}%
    \ifx\svgscale\undefined%
      \relax%
    \else%
      \setlength{\unitlength}{\unitlength * \real{\svgscale}}%
    \fi%
  \else%
    \setlength{\unitlength}{\svgwidth}%
  \fi%
  \global\let\svgwidth\undefined%
  \global\let\svgscale\undefined%
  \makeatother%
  \begin{picture}(1,0.24300546)%
    \lineheight{1}%
    \setlength\tabcolsep{0pt}%
    \put(0,0){\includegraphics[width=\unitlength,page=1]{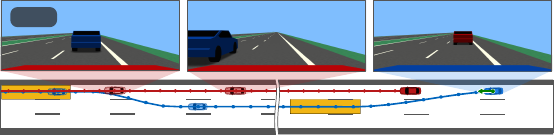}}%
    \put(0.02447347,0.20360169){\color[rgb]{1,1,1}\makebox(0,0)[lt]{\lineheight{1.25}\smash{\begin{tabular}[t]{l}$12.9s$\end{tabular}}}}%
    \put(0,0){\includegraphics[width=\unitlength,page=2]{commonroad_ex.pdf}}%
    \put(0.36162704,0.20340667){\color[rgb]{1,1,1}\makebox(0,0)[lt]{\lineheight{1.25}\smash{\begin{tabular}[t]{l}$12.1s$\end{tabular}}}}%
    \put(0,0){\includegraphics[width=\unitlength,page=3]{commonroad_ex.pdf}}%
    \put(0.70533677,0.20340664){\color[rgb]{1,1,1}\makebox(0,0)[lt]{\lineheight{1.25}\smash{\begin{tabular}[t]{l}$6.9s$\end{tabular}}}}%
  \end{picture}%
\endgroup%

		\caption{{\footnotesize CommonRoad.  The snapshots show the inside view of the vehicle, which is generated by esmini from the OpenSCENARIO file. 
		}}\label{fig:cr-ex1}
	\end{subfigure}
	\vspace{-7mm}
	\caption{Exemplary overtaking scenario.}\label{fig:overtake}
	\vspace{-4.5mm}
\end{figure}

\vspace{1mm}
\textbf{Running example}:
In the scenario\footnote{OpenSCENARIO ID: SimpleOvertake} in Fig.~\ref{fig:osc_ex}, 
an overtaking \texttt{Story} is specified in the \texttt{Storyboard}. To achieve the overtaking task, two instances of the class \texttt{Event} are specified in a single \texttt{Maneuver} object: turn left and turn right. For the first event, vehicle A executes the \texttt{Action} of a lane change to the left when the relative longitudinal distance between the two vehicles falls below $20 m$. For the second event, vehicle A is allowed to perform the lane change to the right when vehicle B is $10 m$ ahead of vehicle A and the time elapsed since the last event is longer than $5 s$.

\subsection{CommonRoad Format}\label{subsec:commonroad}
\begin{figure}[b!]
	\centering 
	\vspace{-5mm}
	\def\svgwidth{0.90\columnwidth}\scriptsize
	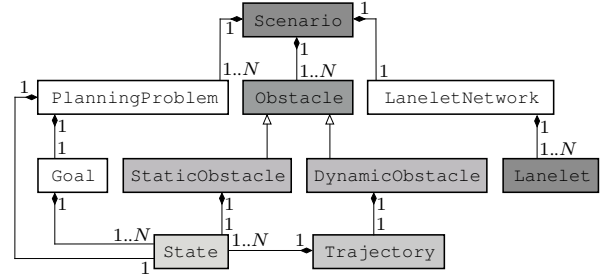
	\caption{UML class diagram of the CommonRoad format. Details of child elements are omitted for clarity.}
		\vspace{-5.5mm}
	\label{fig:cr}
\end{figure}
We present the UML class diagram of the CommonRoad v2023.2 format in Fig.~\ref{fig:cr}. CommonRoad specifies a scenario (class \texttt{Scenario}) with a network of lanenets (class \texttt{LaneletNetwork}), one or several planning problems (class \texttt{PlanningProblem}), and obstacles (class \texttt{Obstacle}). Obstacles are characterized by their role, type (e.g., static or dynamic), shape, and initial state (class \texttt{State}). For dynamic obstacles (class \texttt{DynamicObstacle}), their movement over time is specified by trajectories (class \texttt{Trajectory}) that are a list of states, occupancy sets, or probability distributions. For motion planning, each \textit{ego vehicle}, i.e., the vehicle to be controlled, has an initial state and one or several goal states (class \texttt{Goal}), which are described in the planning problem.

\vspace{1mm}
\textbf{Running example}: An overtaking scenario in CommonRoad format is shown in Fig.~\ref{fig:cr-ex1}. We can observe from the trajectories of both vehicles that vehicle A first changes its lane to the left of vehicle B and then returns to its initial lane. Assuming that vehicle A is the ego vehicle, one can model a \texttt{PlanningProblem} instance by combining its initial state and intermediate goals that are automatically or manually constructed.
\subsection{OpenSCENARIO vs. CommonRoad}\label{subsec:vs}
Both OpenSCENARIO and CommonRoad cover the design and implementation of a traffic scenario, i.e., what should happen and when it is executed. To simulate the \texttt{Storyboard} (cf. Fig.~\ref{fig:openscenario}), OpenSCENARIO requires a director and a simulator core to govern the progress and execute control strategies based on the description. Thus, one cannot easily know how the scenario would look like unless simulating the traffic subject to OpenSCENARIO constraints. Moreover, different vehicle models, control strategies, and computer hardware can all contribute to varying traffic interactions within the simulation.
In contrast, CommonRoad offers two ways to represent scenarios. The first option offers recordings of traffic situations, while the second option offers interactive simulations, i.e., other traffic participants react to the behavior of the ego vehicle through coupling with traffic simulators, such as SUMO \cite{klischat2019coupling}.
On the other hand, OpenSCENARIO itself does not include driver models, vehicle dynamics, and cost functions as in CommonRoad, which currently limits its usage for many applications.

\subsection{Implementation}\label{subsec:imple}
The implementation of our OpenSCENARIO to CommonRoad converter is presented in Alg.~\ref{alg:main}. We begin by creating a lanelet network (see line~\ref{alg:openDrive}) based on the OpenDRIVE file associated with OpenSCENARIO. This is accomplished by calling $\Call{convertOpendriveToLanelets}{}$ described in \cite[Alg.~1]{althoff2018automatic}. Afterwards, a simulation core is used, as described in detail later, to obtain trajectories of all simulated obstacles (see line~\ref{alg:sim}). Then we build a CommonRoad scenario by aggregating the lanelet network and the obstacles (see line~\ref{alg:build_scenario}).  To construct the planning problem, we either employ the predefined ego vehicle in OpenSCENARIO or allow the user to select a vehicle as the ego vehicle (see line~\ref{alg:ego}). In line~\ref{alg:plan}, the planning problem is formulated, e.g., based on the trajectory of the ego vehicle or the scenario descriptions. The conversion ends with writing the CommonRoad scenario in an XML file (see line~\ref{alg:write}).

\renewcommand{\arraystretch}{1.12}
\begin{table}[b!]\small
	\begin{center}
		\vspace{-2.5mm}
		\caption{Conversion of obstacles from OpenSCENARIO to CommonRoad.}
		\label{tab:conversion}\footnotesize
		\begin{tabular}{p{5.cm}p{2.5cm}}
			\toprule[1.1pt]
			\textbf{OpenSCENARIO}  & \textbf{CommonRoad} \\
			\midrule
			\multicolumn{2}{l}{\textbf{Obstacle Type}} \\
			$\mathtt{VEHICLE}.\mathtt{CAR}$, $\mathtt{VEHICLE}.\mathtt{VAN}$& $\mathtt{CAR}$\\
			\makecell[l]{$\mathtt{VEHICLE}.\mathtt{TRUCK}$, $\mathtt{VEHICLE}.\mathtt{TRAILER}$,\\ $\mathtt{VEHICLE}.\mathtt{SEMITRAILER}$} & \multirow{1.}{*}{\centering $\mathtt{TRUCK}$}\\
			$\mathtt{VEHICLE}.\mathtt{BUS}$ & $\mathtt{BUS}$\\
			$\mathtt{VEHICLE}.\mathtt{MOTORBIKE}$ & $\mathtt{MOTOCYCLE}$\\
			$\mathtt{VEHICLE}.\mathtt{BICYCLE}$ & $\mathtt{BICYCLE}$\\
			$\mathtt{VEHICLE}.\mathtt{TRAIN}, \mathtt{VEHICLE}.\mathtt{TRAM}$ & $\mathtt{TRAIN}$\\
			$\mathtt{PEDESTRIAN}$ & $\mathtt{PEDESTRIAN}$\\
			$\mathtt{MISC\_OBJECT.BUILDING}$& $\mathtt{BUILDING}$\\
			$\mathtt{MISC\_OBJECT.TRAFFICISLAND}$& $\mathtt{MEDIAN\_STRIP}$\\
			$\mathtt{MISC\_OBJECT.STREETLAMP}$& $\mathtt{PILLAR}$\\
			\makecell[l]{$\mathtt{MISC\_OBJECT.POLE}$, $\mathtt{MISC\_OBJECT.BARRIER}$,\\ $\mathtt{MISC\_OBJECT.RAILING}$, \\ $\mathtt{MISC\_OBJECT.SOUNDBARRIER}$\vspace{-1mm}}& \multirow{1}{*}{\centering$\mathtt{ROAD\_BOUNDARY}$}\\ 
			$\mathtt{MISC\_OBJECT.PATCH}$& $\mathtt{CONSTRUCTION\_ZONE}$\\
			others & $\mathtt{UNKNOWN}$\\
			\midrule
			\multicolumn{2}{l}{\textbf{State}} \\
			state.timestamp & state.time\_step\\
			$[$state.x, state.y$]$&state.position\\
			state.h&state.orientation\\
			state.speed&state.velocity\\
			state.wheelAngle&state.steering\_angle\\
			state.h\_rate&state.yaw\_rate\\
			\midrule
			\multicolumn{2}{l}{\textbf{Shape}} \\
			state.length&obstacle\_shape.length\\
			state.width&obstacle\_shape.width\\
			\bottomrule[1.0pt]
		\end{tabular}
		\vspace{-2.mm}
	\end{center}
\end{table}

To orchestrate and execute the dynamic elements defined by OpenSCENARIO (cf. Sec.~\ref{subsec:vs}), we utilize esmini as the simulator in line~\ref{alg:sim} of Alg.~\ref{alg:main} because:
\begin{enumerate}
	\item reusing mature software reduces the complexity of the software structure and modularizes the framework;
	\item esmini is more lightweight compared to, e.g., CARLA, yet has relatively high OpenSCENARIO coverage; 
	\item esmini has an interface for SUMO vehicle controllers and can send and receive OSI data, providing flexibility and real-time capabilities for traffic simulation; and
	\item esmini provides a Python interface, which aligns with CommonRoad and many OpenSCENARIO tools.
\end{enumerate}
During the simulation, the states of dynamic obstacles are collected at each frame in the global coordinate system of the converted lanelet network. By default, the esmini simulation ends as soon as all triggered elements (cf. Fig.~\ref{fig:openscenario}) are completed. 
To prevent simulations from running indefinitely due to the absence of \texttt{StopTrigger} elements, we establish an upper time limit $t_{\max}$ for the scenario duration. {Furthermore, we offer the possibility to increase the interactivity of the converter through the UDP interface of esmini, which facilitates the incorporation of external driver models.}

 
After the simulation ends, the esmini Python binding is used to retrieve the information and states of all scenario objects and convert them to CommonRoad types.
Tab.~\ref{tab:conversion} lists the transformation relation from OpenSCENARIO to CommonRoad obstacles. To match the required time step size $\Delta t$ of the CommonRoad scenario, esmini trajectories may need to be resampled, e.g., using methods from \cite[Sec.~3.3]{wang2020coupling}. 

	\section{Numerical Experiments}\label{sec:exp}

\begin{table}[!b]\centering
	\vspace{-2.5mm}\footnotesize
	\caption{Settings and conversion statistics for the conversion.}
	\renewcommand{\arraystretch}{1.0}
	\begin{tabular}{cccc} \toprule
		\multicolumn{2}{c}{\textbf{Parameter}} & \multicolumn{2}{c}{\textbf{Value}} \\ \midrule
		\multicolumn{2}{c}{esmini $\Delta t$} & \multicolumn{2}{c}{$0.01 s$} \\
		\multicolumn{2}{c}{CommonRoad $\Delta t$} & \multicolumn{2}{c}{$0.1 s$}\\
		\multicolumn{2}{c}{$t_{\max}$} & 
		\multicolumn{2}{c}{$60 s$}\\
		\midrule
		\textbf{Source} & \textbf{Success Rate} & \textbf{\makecell{Avg. Conversion\\ Time}} & \textbf{\makecell{Avg. Scenario\\ Duration}}\\\midrule
		I & $100.0\%\ (13/13)$ & $10.68s$ & $53.61s$\\
		II & $100.0 \%\ (26/26)$ & $18.20s$ & $30.96s$\\
		III & $100.0\%\ (15/15)$ & $66.34s$ & $40.64s$\\
		\bottomrule
	\end{tabular}
	\vspace{-1.mm}
	\label{tab:conv_settings}
	
\end{table}
\begin{figure*}[!h]
	\captionsetup[subfigure]{aboveskip=+1mm,belowskip=+1mm}
	\centering
	\begin{subfigure}[b]{0.42\linewidth}
		\footnotesize
		\centering
		\vspace{0.mm}
		\def\svgwidth{0.98\columnwidth}
		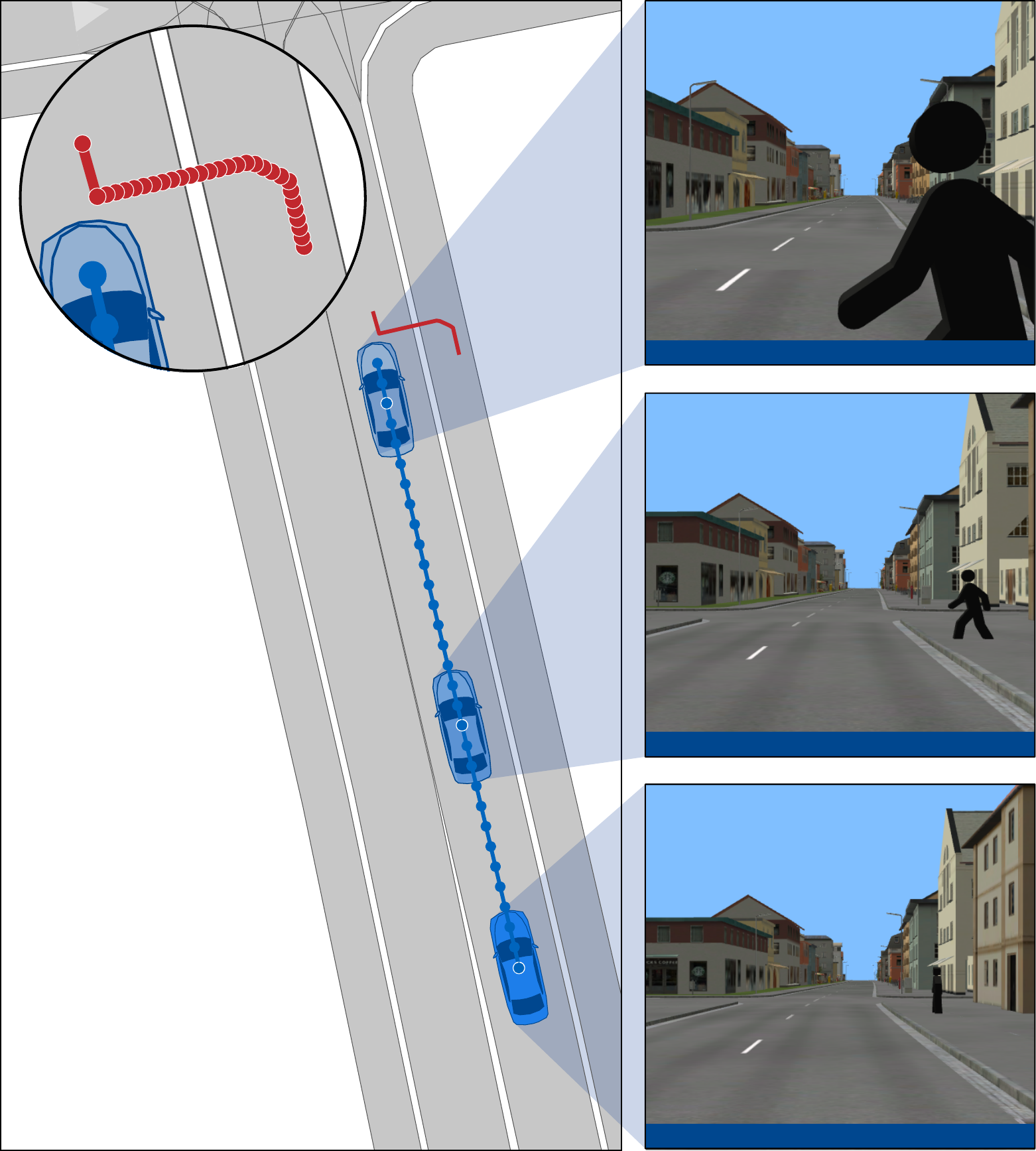
		\caption{{\footnotesize Scenario configuration.}}\label{fig:ped-a}
	\end{subfigure}\hspace{2mm}
	\begin{subfigure}[b]{0.252\linewidth} 
		\footnotesize
		\centering
		\def\svgwidth{0.98\columnwidth}
\begingroup%
  \makeatletter%
  \providecommand\color[2][]{%
    \errmessage{(Inkscape) Color is used for the text in Inkscape, but the package 'color.sty' is not loaded}%
    \renewcommand\color[2][]{}%
  }%
  \providecommand\transparent[1]{%
    \errmessage{(Inkscape) Transparency is used (non-zero) for the text in Inkscape, but the package 'transparent.sty' is not loaded}%
    \renewcommand\transparent[1]{}%
  }%
  \providecommand\rotatebox[2]{#2}%
  \newcommand*\fsize{\dimexpr\f@size pt\relax}%
  \newcommand*\lineheight[1]{\fontsize{\fsize}{#1\fsize}\selectfont}%
  \ifx\svgwidth\undefined%
    \setlength{\unitlength}{450.03899881bp}%
    \ifx\svgscale\undefined%
      \relax%
    \else%
      \setlength{\unitlength}{\unitlength * \real{\svgscale}}%
    \fi%
  \else%
    \setlength{\unitlength}{\svgwidth}%
  \fi%
  \global\let\svgwidth\undefined%
  \global\let\svgscale\undefined%
  \makeatother%
  \begin{picture}(1,1.85000633)%
    \lineheight{1}%
    \setlength\tabcolsep{0pt}%
    \put(0,0){\includegraphics[width=\unitlength,page=1]{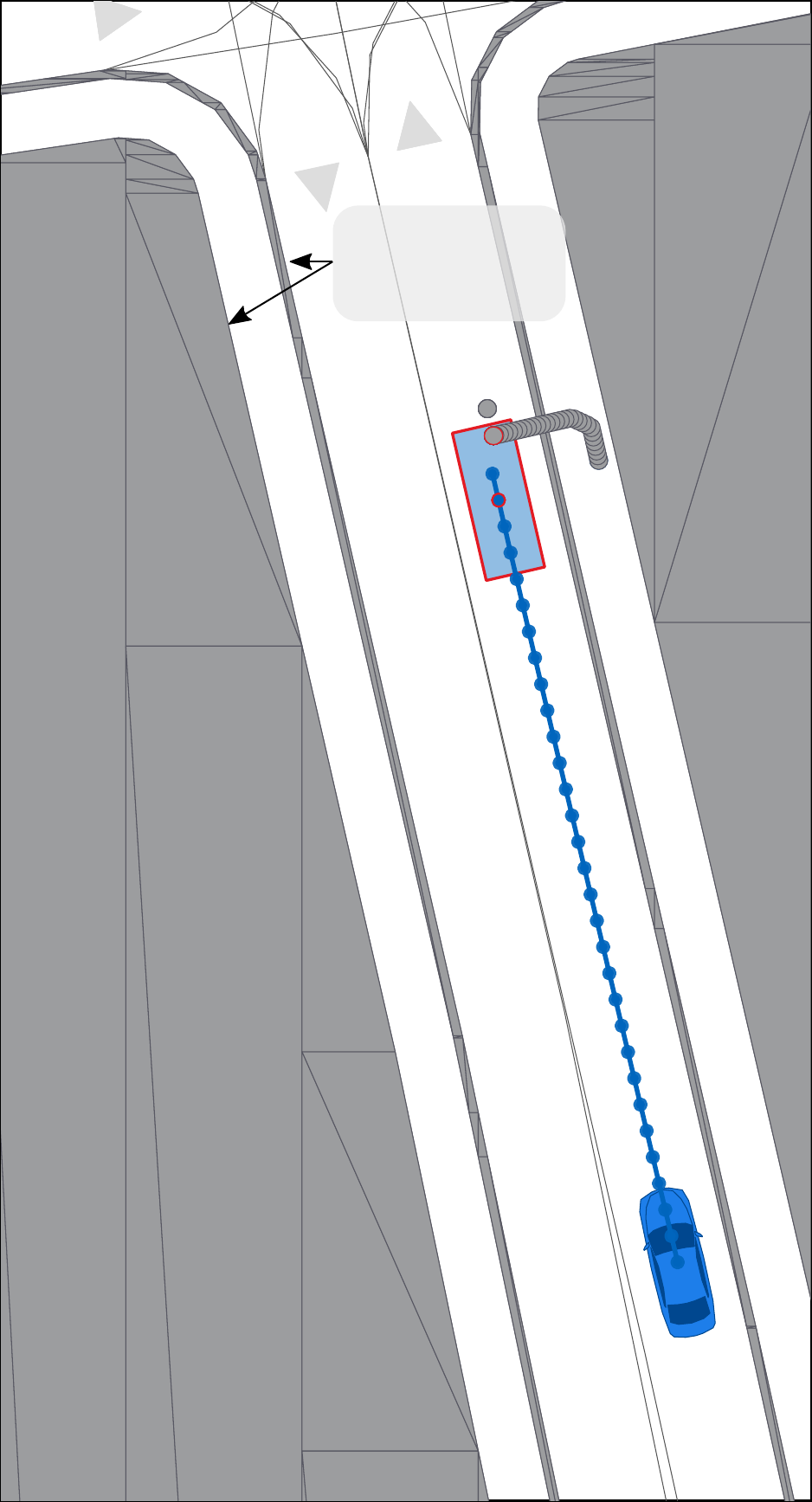}}%
    \put(0.48891942,1.53746198){\color[rgb]{0,0,0}\makebox(0,0)[lt]{\lineheight{1.25}\smash{\begin{tabular}[t]{l}road\end{tabular}}}}%
    \put(0.43776832,1.48156991){\color[rgb]{0,0,0}\makebox(0,0)[lt]{\lineheight{1.25}\smash{\begin{tabular}[t]{l}boundary\end{tabular}}}}%
    \put(0,0){\includegraphics[width=\unitlength,page=2]{collision_checking.pdf}}%
    \put(0.17460263,1.15405171){\color[rgb]{0,0,0}\makebox(0,0)[lt]{\lineheight{1.25}\smash{\begin{tabular}[t]{l}colliding\end{tabular}}}}%
    \put(0.15504649,1.09149358){\color[rgb]{0,0,0}\makebox(0,0)[lt]{\lineheight{1.25}\smash{\begin{tabular}[t]{l}occupancy\end{tabular}}}}%
    \put(0,0){\includegraphics[width=\unitlength,page=3]{collision_checking.pdf}}%
    \put(0.07741618,0.0531631){\color[rgb]{1,1,1}\makebox(0,0)[lt]{\lineheight{1.25}\smash{\begin{tabular}[t]{l}$2.6s$\end{tabular}}}}%
  \end{picture}%
\endgroup%

		\caption{{\footnotesize Collision checking.}}\label{fig:ped-b}
	\end{subfigure}\hspace{2mm}
	\begin{subfigure}[b]{0.252\linewidth} 
		\footnotesize
		\centering
		\def\svgwidth{0.98\columnwidth}
\begingroup%
  \makeatletter%
  \providecommand\color[2][]{%
    \errmessage{(Inkscape) Color is used for the text in Inkscape, but the package 'color.sty' is not loaded}%
    \renewcommand\color[2][]{}%
  }%
  \providecommand\transparent[1]{%
    \errmessage{(Inkscape) Transparency is used (non-zero) for the text in Inkscape, but the package 'transparent.sty' is not loaded}%
    \renewcommand\transparent[1]{}%
  }%
  \providecommand\rotatebox[2]{#2}%
  \newcommand*\fsize{\dimexpr\f@size pt\relax}%
  \newcommand*\lineheight[1]{\fontsize{\fsize}{#1\fsize}\selectfont}%
  \ifx\svgwidth\undefined%
    \setlength{\unitlength}{300.12600021bp}%
    \ifx\svgscale\undefined%
      \relax%
    \else%
      \setlength{\unitlength}{\unitlength * \real{\svgscale}}%
    \fi%
  \else%
    \setlength{\unitlength}{\svgwidth}%
  \fi%
  \global\let\svgwidth\undefined%
  \global\let\svgscale\undefined%
  \makeatother%
  \begin{picture}(1,1.85175507)%
    \lineheight{1}%
    \setlength\tabcolsep{0pt}%
    \put(0,0){\includegraphics[width=\unitlength,page=1]{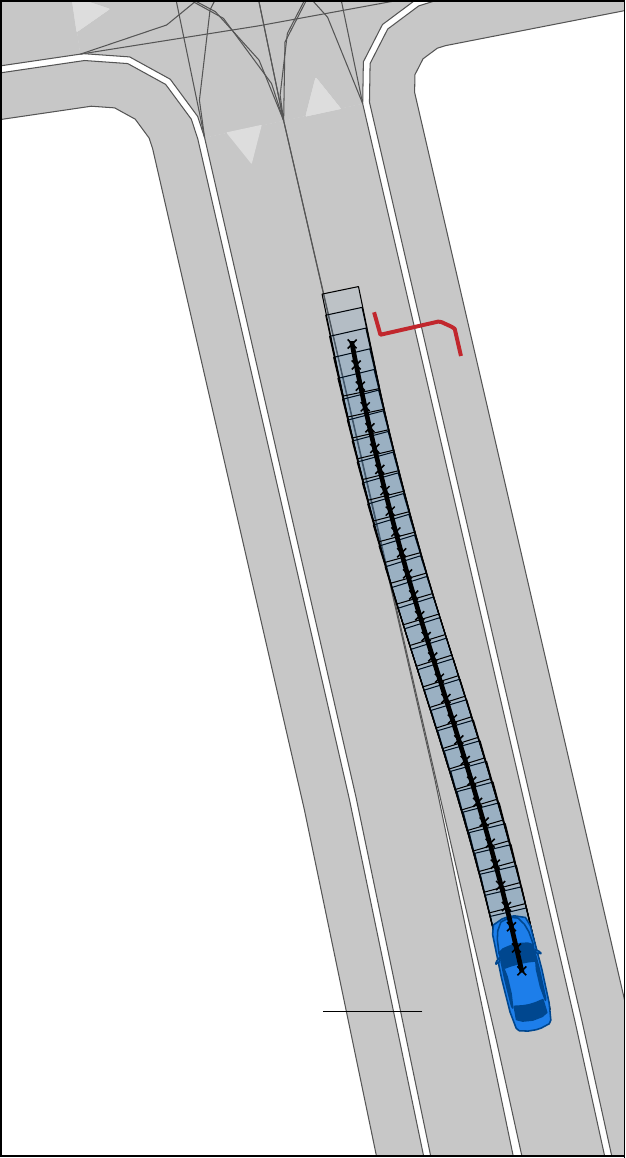}}%
    \put(0.04491485,0.25627904){\color[rgb]{0,0,0}\makebox(0,0)[lt]{\lineheight{1.25}\smash{\begin{tabular}[t]{l}planned trajectory\end{tabular}}}}%
    \put(0.07448576,0.1916124){\color[rgb]{0,0,0}\makebox(0,0)[lt]{\lineheight{1.25}\smash{\begin{tabular}[t]{l}with occupancy\end{tabular}}}}%
    \put(0,0){\includegraphics[width=\unitlength,page=2]{planned.pdf}}%
    \put(0.06740683,1.41573597){\color[rgb]{0,0,0}\makebox(0,0)[lt]{\lineheight{1.25}\smash{\begin{tabular}[t]{l}driving\end{tabular}}}}%
    \put(0.04873181,1.35106932){\color[rgb]{0,0,0}\makebox(0,0)[lt]{\lineheight{1.25}\smash{\begin{tabular}[t]{l}direction\end{tabular}}}}%
    \put(0,0){\includegraphics[width=\unitlength,page=3]{planned.pdf}}%
    \put(0.07905233,0.05322753){\color[rgb]{1,1,1}\makebox(0,0)[lt]{\lineheight{1.25}\smash{\begin{tabular}[t]{l}$2.6s$\end{tabular}}}}%
    \put(0,0){\includegraphics[width=\unitlength,page=4]{planned.pdf}}%
  \end{picture}%
\endgroup%

		\caption{{\footnotesize Motion planning.}}\label{fig:ped-c}
	\end{subfigure}\hspace{-3.0mm}	
	\begin{subfigure}[b]{0.43\linewidth}
		\footnotesize
		\centering
		\def\svgwidth{0.98\columnwidth}
		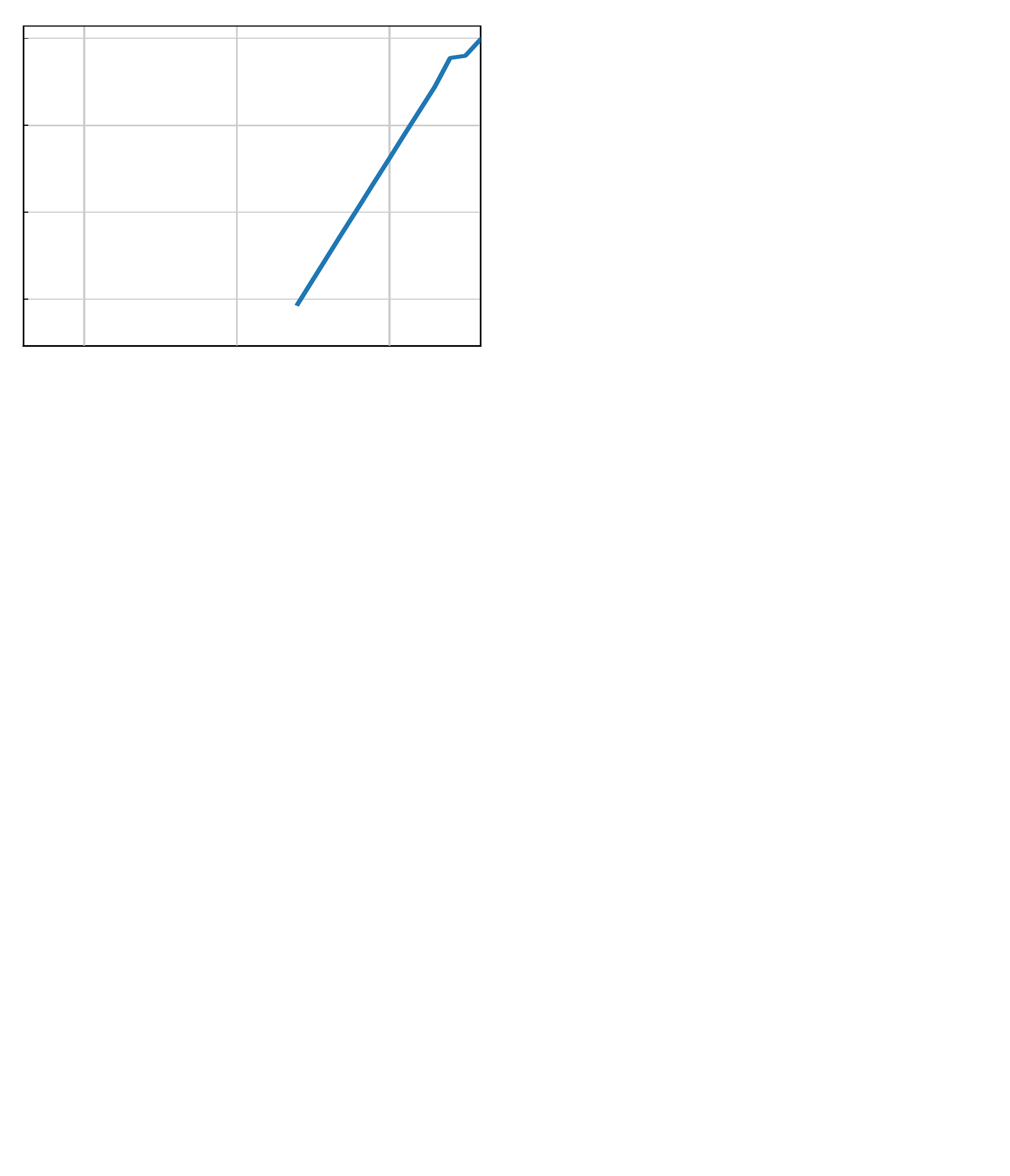\vspace{0mm}
		\caption{{\footnotesize {Criticality measures.}}}\label{fig:ped-d}
	\end{subfigure}\hspace{1.5mm}
	\begin{subfigure}[b]{0.532\linewidth} 
		\footnotesize
		\centering
		\def\svgwidth{0.98\columnwidth}
		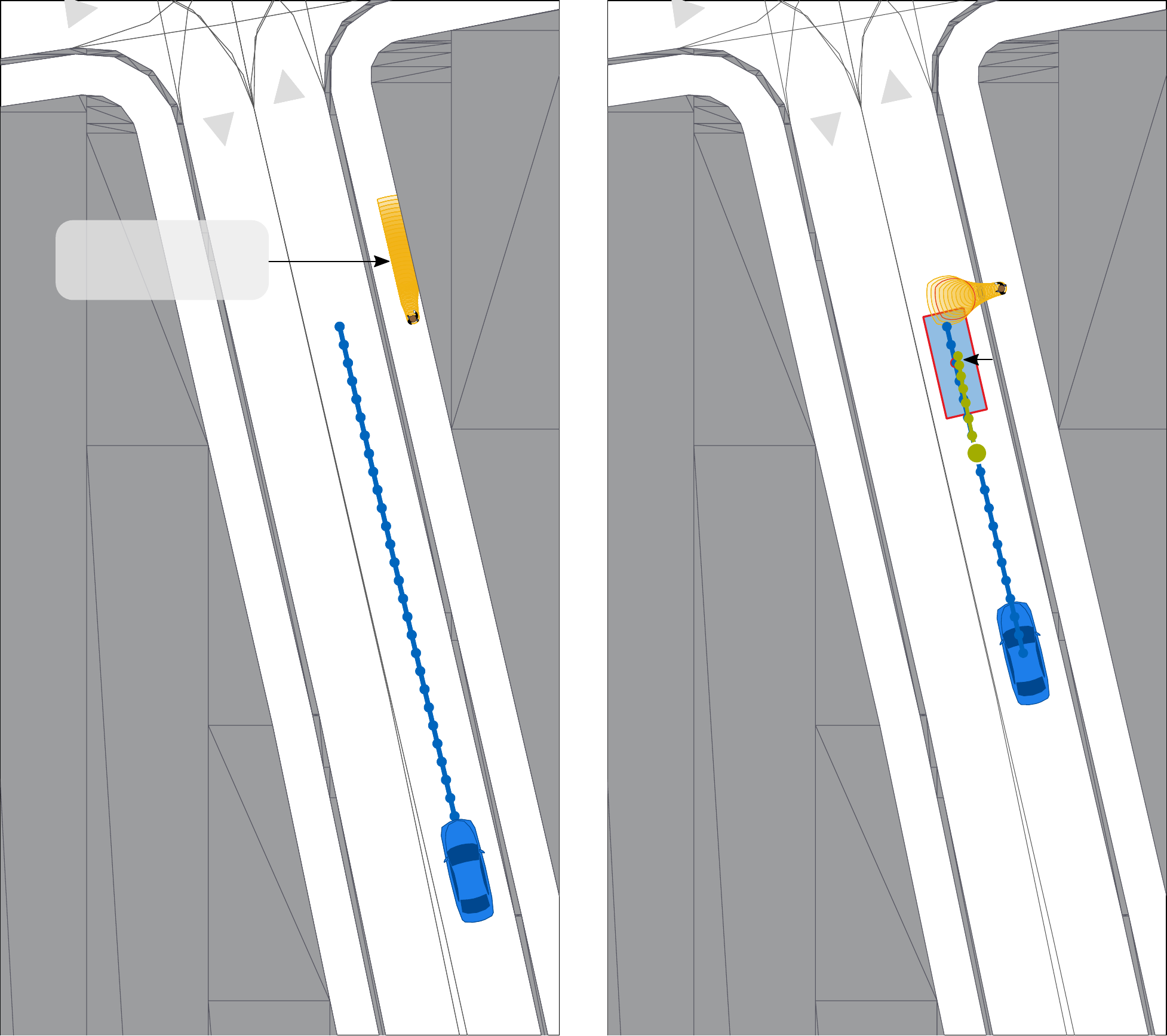\vspace{2.3mm}
		\caption{{\footnotesize Safety verification and trajectory repairing.}}\label{fig:ped-repair}
	\end{subfigure}\vspace{-2.0mm}	
	\caption{{\footnotesize Evaluation results with CommonRoad tools. We only display the scenario information between $2.6s$ and $5.6s$. (a) shows the configuration of the converted CommonRoad scenario, with snapshots captured from the inside view of the ego vehicle during esmini simulation at three time steps. Collision checking and motion planning results are presented in (b) and (c), respectively. To provide clear insights, the criticality of the scenario is plotted on the vertical axis of the graph, with an upward trend indicating increasing criticality, as shown in (d). Finally, (e) displays the safety verification results at both $2.6s$ and $3.8s$, where the trajectory is repaired if the intended trajectory is not legally safe.}}\label{fig:ped}
	\vspace{-5mm} 
\end{figure*}

 To demonstrate the usefulness of our converter, we convert 54 openly accessible OpenSCENARIO scenarios from:
 \begin{enumerate}
 	\item[I.] OpenSCENARIO standard examples,
 	\item[II.] the esmini demonstration package, and
 	\item[III.] automated lane keeping scenarios\footnote{\url{https://github.com/asam-oss/OSC-ALKS-scenarios}}.
 \end{enumerate}
  We exclude OpenSCENARIO files that contain only parameter values and allow certain elements to be reusable. All conversions are performed on a computer with an Intel Core i7-1165G7 CPU and 16 GB of memory. The parameters for the converter are listed in Tab.~\ref{tab:conv_settings}.

\subsection{Conversion Statistics}
The conversion statistics are listed in Tab.~\ref{tab:conv_settings}. Our converter successfully transformed all considered OpenSCENARIO files with an average conversion time of $31.74s$. We observed that the duration of the simulation closely correlates with the complexity of the scenario and the map size. As a result, our converter proves to be an effective tool for converting the OpenSCENARIO description into the CommonRoad format.
\subsection{Scenario Evaluation on CommonRoad}
We demonstrate the practicality of our converter by evaluating an OpenSCENARIO scenario\footnote{OpenSCENARIO ID: pedestrian\_collision} using CommonRoad tools. In this scenario, the ego vehicle and a pedestrian follow predefined routes, where the pedestrian jaywalks. 
Fully comprehending the scenario based solely on the OpenSCENARIO file is challenging. To obtain a deeper understanding, we use our converter to simulate the traffic and show the simulated result in Fig.~\ref{fig:ped-a}. Afterwards, we evaluate the scenario with CommonRoad tools in the following paragraphs:
\paragraph{Collision Checking} we use the open-source toolbox CommonRoad Drivability Checker \cite{PekIV20} to check the drivability of the trajectory of the ego vehicle. It can be verified that this trajectory is kinematically feasible; however, it collides with the pedestrian at $5.5s$. The collision occupancies are highlighted in Fig.~\ref{fig:ped-b}.
\paragraph{Motion Planning} using CommonRoad, motion planners can be easily benchmarked. As shown in Fig.~\ref{fig:ped-c}, the popular motion planner described in \cite{werling2012optimal} successfully avoids collision with the pedestrian by maneuvering the ego vehicle to the left. 
\paragraph{Criticality Comparison} CommonRoad is also equipped with various criticality measures through its tool CommonRoad-CriMe \cite{YuanfeiLin2023CriMe}, which are designed to objectively evaluate the safety and threat level of traffic scenarios. As an example, we use time-to-collision (TTC), worst-time-to-collision (WTTC), time-to-react (TTR), drivable area (DA), brake threat number (BTN), and steer threat number (STN) to evaluate the converted scenario. The curves in Fig.~\ref{fig:ped-d} all show that the scenario is getting more critical over time. The TTR curve indicates that there are no available evasive maneuvers to avoid the collision after $4.8s$.

\paragraph{Safety Verification and Trajectory Repairing} 
CommonRoad offers the tool SPOT \cite{KoschiSPOT} to predict the occupancy set of obstacles based on  legal behaviors. 
	With SPOT, we can safeguard the ego vehicle within a given time interval by ensuring that the planned trajectory is collision-free against the predicted occupancy set \cite{christian2020nature}. As a result, in Fig.~\ref{fig:ped-repair}, the ego vehicle is considered legally safe at $2.6s$ along the intended trajectory but not at $3.8s$, as the latter could pose a danger to the pedestrian when it is inattentive and jaywalks. To efficiently ensure safety for the situation at $3.8s$, we employ the trajectory repairing method described in \cite{YuanfeiLin2021} and \cite{YuanfeiLin2022a}. This approach keeps the trajectory before the TTR unchanged while repairing the remaining part. Thereby the ego vehicle fully brakes to prevent any potential harm to the pedestrian.
	

	\section{Conclusions}\label{sec:con}
	This paper introduces the first publicly available converter from OpenSCENARIO to CommonRoad.	
	By triggering the dynamic elements defined in OpenSCENARIO, the logical scenarios are concretized into CommonRoad format, {incorporating predefined interactions between vehicles}.
	We aim to foster the development, testing, and validation of autonomous driving systems by providing an open-source solution for converting scenarios between different formats. This also serves to bridge the gap between academia and industry, thereby promoting the advancement of technology in the field of autonomous driving.
%

	
	
	\section*{Acknowledgments}
We would like to express our sincere appreciation to Emil Knabe for his invaluable contribution in reviewing and accepting the proposed changes to the esmini interface, and to Sebastian Maierhofer for maintaining the converter from OpenDRIVE to lanelets. Furthermore, the authors gratefully acknowledge partial financial support by the German Federal Ministry for Digital and Transport (BMDV) within the project \textit{Cooperative Autonomous Driving with Safety Guarantees} (KoSi). 
	\bibliographystyle{IEEEtran}
	\balance{\bibliography{openscenario.bib}}
	
\end{document}